\documentclass{article}

\usepackage{arxiv}
\usepackage[utf8]{inputenc} 
\usepackage[T1]{fontenc}    
\usepackage{hyperref}       
\usepackage{url}            
\usepackage{booktabs}       
\usepackage{amsfonts}       
\usepackage{nicefrac}       
\usepackage{microtype}      
\usepackage{lipsum}
\usepackage{graphicx}
\graphicspath{ {./images/} }

\usepackage{amsmath,amsfonts,bm}

\def\eqref#1{equation~\ref{#1}}

\def\1{\bm{1}}

\DeclareMathAlphabet{\mathsfit}{\encodingdefault}{\sfdefault}{m}{sl}
\SetMathAlphabet{\mathsfit}{bold}{\encodingdefault}{\sfdefault}{bx}{n}

\usepackage{natbib}
\usepackage{graphicx}
\usepackage{booktabs}
\usepackage{multirow}
\usepackage{amsmath,amssymb,amsfonts}
\usepackage{amsthm}
\usepackage{mathrsfs}
\usepackage[title]{appendix}
\usepackage{xcolor}
\usepackage{textcomp}
\usepackage{manyfoot}
\usepackage{algorithm}
\usepackage{algorithmicx}
\usepackage{algpseudocode}
\usepackage{listings}
\usepackage{float}
\usepackage{tikz}
\usetikzlibrary{positioning, arrows.meta}
\usepackage{adjustbox,lipsum}
\usepackage{silence}
\ErrorsOff*

\newtheorem{theorem}{Theorem}
\title{A Mean Curvature Approach to Boundary Detection: Geometric Insights for Unsupervised Learning}

\author{
 Alexandre Luis Magalh\~aes Levada\\
  Federal University of S\~ao Carlos\\
  13565-905, S\~ao Carlos-SP, Brazil\\
  \texttt{alexandre.levada@ufscar.br} \\
}

\begin{document}
\maketitle
\begin{abstract}
	Accurate boundary detection in high-dimensional data remains a fundamental challenge in unsupervised learning, particularly in the presence of non-linear structures and heterogeneous densities. In this work, we introduce \textit{Mean Curvature Boundary Points} (MCBP), a novel geometric framework grounded in the emerging field of \textit{Geometric Machine Learning}, which departs from traditional density-based paradigms by explicitly modeling the intrinsic curvature of the data manifold. The proposed method leverages a discrete approximation of the \textit{shape operator}, computed from local $k$-nearest neighbor patches, to estimate pointwise mean curvature without requiring explicit manifold parametrization. The key innovation of MCBP lies in establishing mean curvature as a principled surrogate for boundary characterization: high-curvature regions naturally capture transitions between dense clusters, concave/convex geometric features, and low-density interfaces. This provides a unified geometric interpretation of boundary, outlier, and transition points. In addition, we introduce an adaptive percentile-based thresholding mechanism that enables multiscale boundary extraction, allowing the method to flexibly control boundary thickness without relying on ad hoc density parameters. Beyond detection, we propose a curvature-driven data decomposition strategy that separates samples into smooth (low-curvature) and boundary (high-curvature) subsets, effectively acting as a non-linear geometric filter. This decomposition enhances cluster separability and improves the stability of downstream unsupervised algorithms. Extensive experiments on synthetic and real-world datasets demonstrate that MCBP consistently improves clustering performance when used as a preprocessing step, outperforming classical approaches in scenarios involving complex geometries and high dimensionality. Our results position MCBP as a concrete contribution to Geometric Machine Learning, highlighting the potential of curvature-aware analysis as a unifying paradigm that bridges differential geometry and data-driven modeling, and opening new avenues for incorporating geometric operators into modern machine learning pipelines.
\end{abstract}

\section{Introduction}\label{sec1}

\noindent
Detecting boundary points in multivariate datasets is a fundamental problem in machine learning, as these samples encode the interfaces separating distinct regions of the data space, which may correspond to different classes or clusters \cite{Boundary3,Boundary6}. In high-dimensional settings, however, this task becomes particularly challenging due to phenomena such as distance concentration, sparsity, and the degradation of local neighborhood structures, which hinder reliable estimation of proximity and density \cite{beyer1999nearest, bellman1961curse}. As a result, accurately identifying boundary points, typically located at the intersection of multiple underlying distributions, remains a non-trivial problem, further exacerbated by their intrinsic ambiguity and sensitivity to noise \cite{Boundary1,Boundary2}.

Most existing approaches to boundary detection rely predominantly on density-based principles, where samples in low-density regions are labeled as boundary or noise, while dense regions define cluster cores \cite{ester1996dbscan, breunig2000lof, campello2015hdbscan}. Although effective in well-separated and homogeneous datasets, such methods implicitly assume that density variations alone are sufficient to characterize the underlying structure of the data. This assumption is often violated in practice, particularly in scenarios involving non-linear manifolds, anisotropic distributions, or heterogeneous sampling densities, where geometric properties such as curvature and local shape become essential descriptors \cite{donoho2003hessian, coifman2006diffusion}. Consequently, purely density-driven approaches may fail to distinguish between flat low-density regions and genuinely high-curvature transition zones, leading to imprecise or unstable boundary identification.

Despite these challenges, boundary points play a central role across both supervised and unsupervised learning tasks, as they implicitly determine the geometry of decision regions and the separation between clusters. In classification problems, boundary points directly influence the estimation of decision boundaries, which partition the feature space into class-specific regions; their accurate identification is particularly critical in the presence of non-linear or complex class structures \cite{Boundary4}. Mischaracterizing such points can lead to poorly positioned decision boundaries and degraded generalization performance, a phenomenon especially evident in margin-based models such as support vector machines. In clustering, boundary points define the interfaces between groups and are essential for capturing the true shape and extent of each cluster \cite{Boundary5}. This is particularly relevant in density-based methods such as DBSCAN and HDBSCAN \cite{DBSCAN,HDBSCAN1,HDBSCAN2}, where distinguishing between core, boundary, and noise points directly impacts the quality of the resulting partition. Failure to properly identify boundary samples may lead to clusters that are either overly fragmented or excessively merged, ultimately resulting in underfitting or overfitting of the underlying data structure.

From a broader perspective, isolating boundary regions can be interpreted as a principled preprocessing strategy that separates high-uncertainty samples from stable core regions, reducing noise and enhancing the robustness of downstream learning algorithms. This often translates into improved cluster coherence, more accurate decision boundaries, and higher-quality low-dimensional representations \cite{ARDOD,overfitting}. 

These observations expose a critical gap in the current literature: while boundary detection has been traditionally framed as a density estimation problem, there is a lack of methods that explicitly incorporate intrinsic geometric information into this process. Addressing this limitation is essential for advancing boundary detection in complex, high-dimensional datasets, and aligns with the broader perspective of Geometric Machine Learning, which emphasizes the role of manifold structure and differential operators in data analysis \cite{bronstein2017geometric,GML,Papillon2025}.

To address these limitations, we propose \textit{Mean Curvature Boundary Points} (MCBP), a novel boundary detection framework grounded in principles of differential geometry and aligned with the emerging paradigm of Geometric Machine Learning. Instead of relying solely on density estimators, MCBP explicitly models the intrinsic geometry of the data by approximating the local \textit{shape operator} on $k$-nearest neighbor patches, enabling the estimation of pointwise mean curvature without requiring an explicit manifold parametrization. This formulation introduces a fundamental shift in perspective: boundary points are characterized as regions of high curvature, capturing not only low-density areas but also geometrically complex transitions such as concave and convex interfaces that are typically overlooked by classical methods.

The proposed approach offers three main contributions. First, it provides a principled geometric criterion for boundary detection based on mean curvature, unifying the identification of outliers, transition regions, and decision interfaces under a single mathematical framework. Second, it incorporates an adaptive percentile-based thresholding mechanism that enables multiscale boundary extraction, allowing the method to flexibly control the thickness of the detected boundary without introducing sensitive density parameters. Third, we introduce a curvature-driven data decomposition strategy that partitions the dataset into smooth (low-curvature) and boundary (high-curvature) components, effectively acting as a non-linear geometric filter that enhances cluster separability and improves the stability of downstream learning algorithms.

In contrast to traditional density-based approaches, which may fail in the presence of non-linear manifolds or heterogeneous sampling, MCBP leverages intrinsic geometric information to provide a more robust and expressive characterization of data structure. This enables more accurate boundary localization in high-dimensional and complex datasets, while naturally connecting boundary detection with curvature-aware representations, thereby bridging classical unsupervised learning techniques with modern geometric perspectives.

Empirical results on clustering tasks indicate that the proposed MCBP framework can be effectively employed as a preprocessing step in unsupervised learning, yielding consistent improvements in both centroid initialization and overall clustering quality. By filtering out high-curvature boundary points, the method produces a smoother representation of the data in which the intrinsic cluster structure becomes more prominent, facilitating the identification of representative centroids located in high-density regions. This, in turn, leads to more stable and reliable initialization strategies, particularly for algorithms such as $k$-means, which are known to be sensitive to centroid placement. Furthermore, the removal of geometrically ambiguous samples reduces the impact of noise and overlapping regions, resulting in clusters that are more compact, better separated, and more consistent with the underlying data distribution. These findings suggest that curvature-driven preprocessing not only enhances the robustness of clustering algorithms but also provides a principled mechanism for improving their performance in complex, high-dimensional settings.

The remaining of the paper is organized as follows: Section 2 presents the related work, focusing in recent relevant works on the subject. Section 3 describes the proposed MCBP algorithm in details, explaining the fundamental concepts and intuition behind the method. Section 4 presents the computational experiments and the obtained results. Finally, Section 5 shows our conclusions and final remarks.

\section{Related Work}\label{sec2}

Boundary point detection has emerged as an important component in unsupervised learning, particularly in clustering and outlier analysis, where identifying transition regions between data groups is essential for accurate structure recovery. Early approaches are predominantly rooted in density-based principles, where boundary points are inferred from local variations in neighborhood density. A representative example is BRIM \cite{Brim}, which estimates a \emph{boundary degree} based on the asymmetry between positive and negative $\epsilon$-neighborhoods. By exploiting local distributional properties, BRIM improves upon earlier methods such as DBSCAN \cite{DBSCAN} and BORDER \cite{Boundary3}, achieving higher efficiency and robustness in noisy settings. However, like most density-driven approaches, its performance is sensitive to parameter choices (e.g., $\delta$), which directly influence the number of detected boundary points and limit its adaptability across datasets with heterogeneous structures.

Subsequent works have sought to mitigate these limitations by incorporating alternative structural cues. For instance, the Boundary Point Detection (BPD) algorithm \citep{graphics} avoids explicit thresholding and focuses on accurately capturing concave and convex edges in 3D point clouds, complemented by the Raw Boundary Smoothing (RBS) method, which applies FFT-based filtering to improve boundary regularity. While effective in geometric reconstruction tasks, these approaches are primarily tailored to low-dimensional spatial data and do not readily generalize to high-dimensional machine learning scenarios.

To address high-dimensional challenges, more sophisticated representations have been proposed. The method in \citep{Markov} introduces a directed Markov tree structure to model local feature dependencies, enabling boundary detection via a combinatorial traversal scheme. Although this approach improves performance in domains such as gene expression analysis, its reliance on structured projections and discrete traversal strategies limits its interpretability from a geometric standpoint. In parallel, the Boundary Point Factor (BPF) method \citep{Boundary4}, inspired by the Local Outlier Factor (LOF) \cite{LOF}, integrates density and distance information into a unified scoring function. BPF is notable for requiring only a single parameter and for its robustness to varying densities and outliers. Its extensions, StaticBPF and StreamBPF \citep{Boundary5}, further improve scalability and enable real-time processing through incremental updates and grid-based neighbor search. Despite these advances, such methods remain fundamentally tied to density estimation and do not explicitly account for intrinsic geometric properties of the data manifold.

Beyond clustering, boundary-aware representations have also been explored in application-driven contexts. For example, BPG3D \citep{drive} leverages boundary points to enhance 3D object detection in LiDAR point clouds, introducing specialized modules such as Boundary Point Pooling (BPP) and feature fusion mechanisms. While demonstrating strong empirical performance in computer vision benchmarks, these approaches are tightly coupled to domain-specific architectures and do not provide a general framework for boundary detection in abstract feature spaces.

Finally, methods such as ARDOD \cite{ARDOD} attempt to improve robustness by combining density and distance measures in a parameter-free setting, dynamically adapting to the underlying data distribution. Although effective for outlier detection, ARDOD and related techniques still rely on extrinsic notions of proximity and density, which may be insufficient to capture complex geometric phenomena such as curvature and manifold anisotropy.

Overall, the existing literature is largely dominated by density-driven or hybrid approaches, which, despite their empirical success, exhibit fundamental limitations when dealing with non-linear, high-dimensional, or geometrically complex datasets. In particular, there is a lack of methods that explicitly incorporate intrinsic geometric information, such as curvature, into the boundary detection process. This gap motivates the development of approaches grounded in differential geometry, as pursued in this work.

Overall, the existing literature is largely dominated by density-driven or hybrid approaches, which, despite their empirical success, exhibit fundamental limitations when dealing with non-linear, high-dimensional, or geometrically complex datasets. In particular, there is a lack of methods that explicitly incorporate intrinsic geometric information, such as curvature, into the boundary detection process. This gap motivates the development of approaches grounded in differential geometry, as pursued in this work. In the following section, we introduce the proposed Mean Curvature Boundary Points (MCBP) framework, which estimates local mean curvature via discrete approximations of the shape operator on $k$-nearest neighbor graphs, providing a principled and geometry-aware formulation for boundary detection.

\section{Differential Geometry Basics}

This section provides a concise overview of fundamental concepts from differential geometry that underpin the proposed method. In particular, we introduce the notions of tangent space, the first fundamental form (or metric tensor), the second fundamental form, the shape operator, and mean curvature, key objects for characterizing the local geometry of smooth manifolds. These concepts offer a principled framework for describing how a manifold bends and stretches in its ambient space, enabling the analysis of intrinsic and extrinsic geometric properties \cite{spivak1999comprehensive, ONeill2006, DGApp, do_carmo_differential_2016, tu2017differential, needham2021visual}. In the context of data analysis, they provide the mathematical foundation for modeling datasets as discrete samples from an underlying manifold and for quantifying local geometric variations through differential operators. The following exposition is tailored to establish the necessary intuition and formalism required to derive our curvature-based boundary detection approach.

\subsection{Tangent Spaces}
\noindent
Let $\mathcal{M} \subset \mathbb{R}^m$ be a smooth $d$-dimensional manifold embedded in an ambient space. The \emph{tangent space} at a point $\mathbf{x} \in \mathcal{M}$, denoted by $T_{\mathbf{x}}\mathcal{M}$, is a $d$-dimensional vector space that locally approximates the manifold around $\mathbf{x}$. Formally, $T_{\mathbf{x}}\mathcal{M}$ can be defined as the span of the partial derivatives of a smooth parametrization $\boldsymbol{\phi}: U \subset \mathbb{R}^d \rightarrow \mathcal{M}$ evaluated at $\mathbf{x} = \boldsymbol{\phi}(\mathbf{u})$, that is,
\[
T_{\mathbf{x}}\mathcal{M} = \mathrm{span} \left\{ \frac{\partial \boldsymbol{\phi}}{\partial u_1}, \dots, \frac{\partial \boldsymbol{\phi}}{\partial u_d} \right\}.
\]
Equivalently, it can be characterized as the set of velocity vectors of all smooth curves $\boldsymbol{\gamma}(t)$ lying on the manifold such that $\boldsymbol{\gamma}(0) = \mathbf{x}$, i.e.,
\[
T_{\mathbf{x}}\mathcal{M} = \left\{ \boldsymbol{\gamma}'(0) \;:\; \boldsymbol{\gamma}(t) \in \mathcal{M},\ \boldsymbol{\gamma}(0) = \mathbf{x} \right\}.
\]

Intuitively, the tangent space can be understood as the best linear approximation of the manifold in a small neighborhood around $\mathbf{x}$. While the manifold itself may exhibit complex non-linear structure globally, at sufficiently small scales it behaves approximately like a flat Euclidean space. The tangent space captures this local linearity, acting as a plane (or hyperplane) that ``touches'' the manifold at $\mathbf{x}$ and aligns with its local direction of variation. For example, for a two-dimensional surface embedded in $\mathbb{R}^3$, the tangent space corresponds to the familiar tangent plane at a point on the surface.

From a data analysis perspective, the tangent space plays a central role in manifold-based learning, as it provides a local coordinate system in which geometric quantities can be approximated from discrete samples. In practice, when only a finite set of data points is available, the tangent space at a sample can be estimated from its local neighborhood, typically using techniques such as Principal Component Analysis (PCA), which identifies the directions of maximum variance and approximates the underlying manifold directions. This local linear structure serves as the foundation for defining more advanced geometric objects, such as metric tensors and curvature operators, which are essential for the proposed method.

\begin{figure}
	\centering
	\includegraphics[scale=0.2]{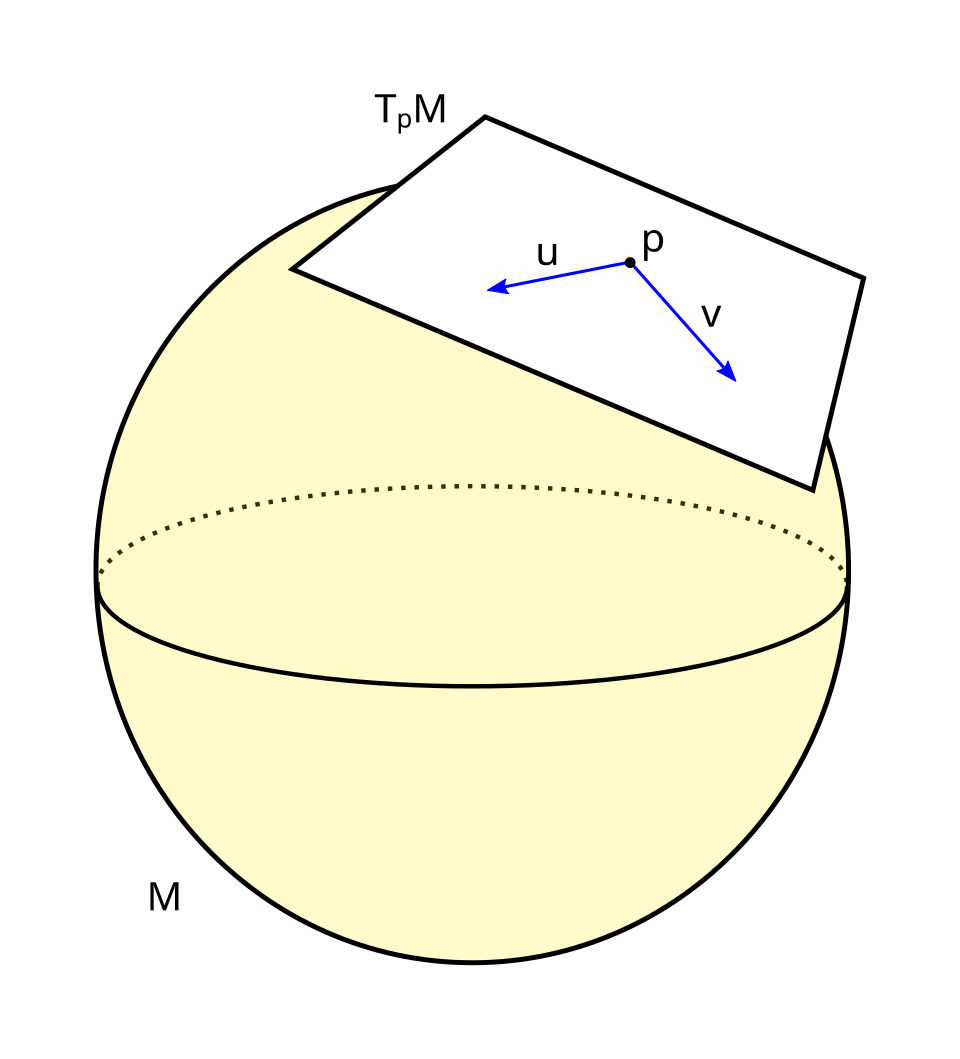}
	\caption{Illustration of the tangent space at a point $\mathbf{x}$ on a sphere embedded in $\mathbb{R}^3$. The tangent plane $T_{\mathbf{x}}\mathcal{M}$ provides a first-order linear approximation of the manifold in a local neighborhood of $\mathbf{x}$, capturing the directions of admissible infinitesimal variations along the surface.}
	\label{fig:tangent}
\end{figure}

Figure \ref{fig:tangent} illustrates the tangent plane to a sphere at a given point. Within the framework of geometric machine learning, tangent spaces are essential because they allow complex nonlinear structures to be approximated by locally linear models. A wide range of algorithms, including manifold learning techniques and graph-based methods, implicitly rely on the assumption that, in sufficiently small neighborhoods, the data lies close to a linear subspace. This perspective motivates the use of methods such as local PCA to obtain empirical estimates of $T_{\mathbf{x}}\mathcal{M}$ from sampled data.

From a geometric standpoint, tangent spaces encode first-order information about the manifold, enabling the definition of local coordinate systems, projection of data points, and approximation of geodesic distances. Furthermore, they serve as the basis for introducing higher-order geometric descriptors, such as curvature, which quantify deviations from this linear approximation. In contemporary learning frameworks, accurate estimation of tangent spaces is crucial for building representations that preserve local structure while supporting global inference, particularly in applications involving dimensionality reduction, semi-supervised learning, and metric learning in non-Euclidean settings.

\subsection{First Fundamental Form (Metric Tensor)}
\noindent
The \emph{first fundamental form}, also known as the \emph{metric tensor}, provides a formal mechanism for measuring geometric quantities, such as lengths, angles, and inner products, on a manifold. Let $\mathcal{M} \subset \mathbb{R}^m$ be a smooth $d$-dimensional manifold with a local parametrization $\boldsymbol{\phi}: U \subset \mathbb{R}^d \rightarrow \mathcal{M}$. The metric tensor at a point $\mathbf{x} = \boldsymbol{\phi}(\mathbf{u})$ is defined as the matrix
\[
\mathbf{G}(\mathbf{u}) = \left[ \left\langle \frac{\partial \boldsymbol{\phi}}{\partial u_i}, \frac{\partial \boldsymbol{\phi}}{\partial u_j} \right\rangle \right]_{i,j=1}^d,
\]
where $\langle \cdot, \cdot \rangle$ denotes the standard inner product in $\mathbb{R}^m$. This matrix encodes how the local coordinate system induced by the parametrization distorts distances and angles relative to the ambient space. Equivalently, the first fundamental form defines a quadratic form that measures the squared length of an infinitesimal displacement $d\mathbf{u} \in \mathbb{R}^d$ on the manifold:
\[
ds^2 = d\mathbf{u}^\top \mathbf{G}(\mathbf{u}) \, d\mathbf{u}.
\]
This expression generalizes the notion of Euclidean distance to curved spaces, allowing one to compute intrinsic geometric quantities directly on the manifold.

From an intuitive perspective, the metric tensor can be understood as a local “ruler” that adapts to the geometry of the data. While Euclidean space assumes uniform scaling in all directions, the metric tensor captures anisotropies and local deformations, indicating how distances are stretched or compressed along different directions of the tangent space. In particular, directions associated with larger metric coefficients correspond to regions where small changes in coordinates produce larger displacements in the ambient space.

In the context of data analysis, the metric tensor plays a central role in modeling the local structure of datasets that lie on or near a manifold. When only discrete samples are available, it is common to approximate the metric tensor using statistics of local neighborhoods. A widely used approach consists of estimating the covariance matrix of points in a $k$-nearest neighbor patch, which captures the principal directions of variation and provides a data-driven approximation of the local geometry. This connection between covariance structure and metric estimation forms the basis for many manifold learning and metric learning techniques, and it is a key component of the curvature-based framework proposed in this work.

\subsection{Second Fundamental Form}

\noindent
While the first fundamental form captures intrinsic properties of a manifold, the \emph{second fundamental form} characterizes how the manifold bends within the ambient space. Let $\mathcal{M} \subset \mathbb{R}^m$ be a smooth $d$-dimensional manifold with local parametrization $\boldsymbol{\phi}: U \subset \mathbb{R}^d \rightarrow \mathcal{M}$ and let $\mathbf{x} = \boldsymbol{\phi}(\mathbf{u})$. The second fundamental form at $\mathbf{x}$ is defined in terms of the second-order derivatives of the parametrization, projected onto the normal space of the manifold. More precisely, if $\mathbf{n}$ denotes a unit normal vector at $\mathbf{x}$ (for hypersurfaces), the coefficients of the second fundamental form are given by
\[
\mathbf{B}_{ij} = \left\langle \frac{\partial^2 \boldsymbol{\phi}}{\partial u_i \partial u_j}, \mathbf{n} \right\rangle,
\]
and the second fundamental form can be written as the quadratic form
\[
\mathrm{II}(d\mathbf{u}) = d\mathbf{u}^\top \mathbf{B} \, d\mathbf{u}.
\]

Geometrically, the second fundamental form measures how the tangent space changes as one moves along the manifold. While the tangent space provides a first-order (linear) approximation, the second fundamental form captures second-order effects, describing how the manifold deviates from this local linear model. In particular, it quantifies the rate at which the normal direction varies, providing a direct measure of local curvature.

An intuitive way to understand this concept is to consider a surface embedded in $\mathbb{R}^3$. At a given point, the tangent plane approximates the surface locally. However, as one moves away from that point, the surface may bend upward, downward, or remain flat. The second fundamental form encodes this bending behavior: directions in which the surface curves strongly correspond to larger values of $\mathrm{II}$, while flat directions yield values close to zero. Thus, it distinguishes between locally flat regions and regions of significant geometric variation.

In the context of data analysis, the second fundamental form provides a mechanism to capture curvature information from sampled data. Although the underlying manifold is not explicitly known, second-order structure can be approximated from local neighborhoods using techniques based on higher-order statistics or Hessian operators. This enables the estimation of curvature-related quantities, which are essential for identifying regions of geometric complexity, such as boundaries, beyond what can be inferred from first-order (metric) information alone.

\subsection{Shape Operator}

\noindent
The \emph{shape operator} (also known as the \emph{Weingarten map}) provides a linear operator that links intrinsic and extrinsic geometry by describing how the normal direction to a manifold varies along tangent directions. Let $\mathcal{M} \subset \mathbb{R}^m$ be a smooth $d$-dimensional manifold and let $\mathbf{x} \in \mathcal{M}$. For hypersurfaces with a unit normal vector field $\mathbf{n}(\mathbf{x})$, the shape operator at $\mathbf{x}$ is defined as
\[
\mathbf{S}_{\mathbf{x}} : T_{\mathbf{x}}\mathcal{M} \rightarrow T_{\mathbf{x}}\mathcal{M}, 
\quad \mathbf{S}_{\mathbf{x}}(\mathbf{v}) = - D\mathbf{n}(\mathbf{x})[\mathbf{v}],
\]
where $D\mathbf{n}(\mathbf{x})$ denotes the differential of the normal field. Intuitively, $\mathbf{S}_{\mathbf{x}}$ measures how the normal vector changes when moving in a tangent direction $\mathbf{v}$. An equivalent and often more practical formulation expresses the shape operator in terms of the first and second fundamental forms. In matrix form, it can be written as
\[
\mathbf{S} = \mathbf{G}^{-1}\mathbf{B},
\]
where $\mathbf{G}$ is the metric tensor (first fundamental form) and $\mathbf{B}$ is the matrix associated with the second fundamental form. This relation highlights that the shape operator can be interpreted as a normalized curvature tensor, combining first-order (metric) and second-order (curvature) information.

From a geometric perspective, the eigenvalues of $\mathbf{S}$ correspond to the \emph{principal curvatures}, and the associated eigenvectors define the principal directions along which the manifold bends the most or the least. A particularly important scalar quantity derived from the shape operator is the \emph{mean curvature}, defined as the trace of $\mathbf{S}$:
\[
H(\mathbf{x}) = \mathrm{tr}(\mathbf{S}_{\mathbf{x}}).
\]
Mean curvature provides a compact summary of local bending, aggregating curvature contributions across all tangent directions. Regions with high mean curvature typically correspond to sharp geometric transitions, while low values indicate locally flat or smooth areas.

Intuitively, the shape operator can be seen as a device that measures how much the manifold “turns” as one moves along different directions. While the tangent space captures the best linear approximation of the manifold, the shape operator quantifies how this approximation changes, effectively encoding second-order geometric behavior. For example, on a sphere, the normal vector changes uniformly in all directions, leading to constant positive curvature; in contrast, on a flat plane, the normal vector remains constant, yielding zero curvature.

In the context of Geometric Machine Learning, the shape operator plays a central role as it provides a principled way to extract curvature information from data. By combining local estimates of the metric tensor and second-order structure (e.g., via covariance and Hessian approximations), one can construct discrete analogues of $\mathbf{S}$ directly from sampled datasets. This enables the computation of curvature-based descriptors, such as mean curvature, which are highly informative for identifying regions of geometric complexity. In particular, boundary points, transition regions, and outliers often manifest as areas of high curvature, making the shape operator a powerful tool for tasks such as boundary detection, manifold learning, and structure-aware data analysis.

\begin{figure}
	\centering
	\includegraphics[scale=0.6]{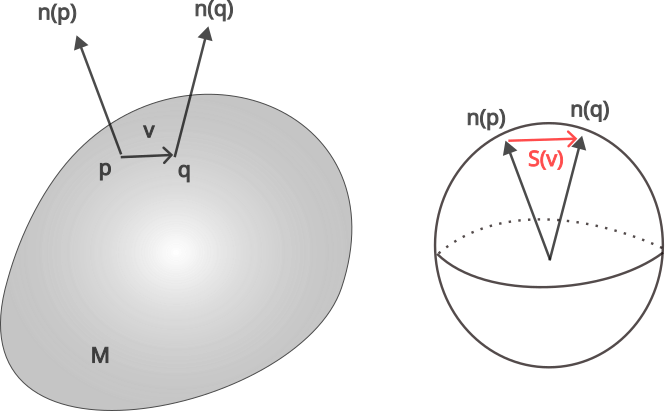}
	\caption{The shape operator quantifies the variation of the normal vector field along a tangent direction $\vec{v}$. More precisely, it measures the rate at which the surface normal changes as one moves from the base point in the direction of $\vec{v}$, providing a local characterization of how the surface bends in different directions.}
	\label{fig:shape}
\end{figure}

Figure~\ref{fig:shape} provides a geometric illustration of the shape operator, highlighting its role in capturing the directional variation of the normal vector across a curved surface. In particular, it shows how the rate of change of the normal vector depends on the chosen tangent direction, thereby encoding the anisotropic nature of local curvature.

\section{The Mean Curvature Boundary Points Algorithm}
\label{sec:MCBP}

\noindent
Building upon the geometric foundations introduced in the previous section, we now present the \textit{Mean Curvature Boundary Points} (MCBP) algorithm, a novel framework for boundary detection grounded in differential geometry. The central premise of MCBP is that boundary points can be effectively characterized as regions of high mean curvature, reflecting abrupt geometric transitions in the underlying data manifold. To operationalize this idea in the context of discrete datasets, we develop a computational pipeline that approximates local tangent spaces, estimates first- and second-order geometric structure from $k$-nearest neighbor neighborhoods, and constructs a discrete analogue of the shape operator. This enables the estimation of pointwise mean curvature without requiring explicit manifold parametrization. The resulting curvature signal is then leveraged to identify boundary regions through an adaptive, data-driven thresholding strategy. In contrast to traditional density-based methods, the proposed approach directly exploits intrinsic geometric information, providing a more expressive and robust characterization of complex, high-dimensional data structures.

\noindent
Let $X = [\mathbf{x}_1, \ldots, \mathbf{x}_n] \in \mathbb{R}^{m \times n}$ denote a dataset composed of $n$ samples embedded in an $m$-dimensional feature space, where each column $\mathbf{x}_i \in \mathbb{R}^m$ corresponds to an individual observation. In order to capture the local geometric structure of the data, we construct a $k$-nearest neighbor graph ($k$-NNG), a widely used tool in manifold learning and graph-based methods \citep{KNNG1}. Specifically, for each sample $\mathbf{x}_i$, we identify its set of $k$ nearest neighbors, denoted by $\eta_i = \{\mathbf{x}_{i1}, \ldots, \mathbf{x}_{ik}\}$, according to a chosen distance metric, typically the Euclidean distance in the ambient space.\footnote{Other metrics may be employed depending on the application domain or feature representation.} The dataset is then represented as a graph whose nodes correspond to samples and whose edges encode local proximity relationships by connecting each point to its $k$ nearest neighbors.

Based on this construction, we define a local neighborhood (or \emph{patch}) around each sample $\mathbf{x}_i$ as the set $P_i = \{\mathbf{x}_i\} \cup \eta_i$, which contains $k+1$ points. This patch provides a discrete approximation of the local manifold structure in the vicinity of $\mathbf{x}_i$ and serves as the fundamental unit for estimating geometric quantities. In matrix form, the patch $P_i$ can be expressed as
\begin{align}
	P_i =
	\begin{bmatrix}
		x_i(1) & x_{i1}(1) & \cdots & x_{ik}(1) \\
		x_i(2) & x_{i1}(2) & \cdots & x_{ik}(2) \\
		\vdots & \vdots    & \ddots & \vdots    \\
		x_i(m) & x_{i1}(m) & \cdots & x_{ik}(m)
	\end{bmatrix}
	\in \mathbb{R}^{m \times (k+1)}.
\end{align}
These local patches play a central role in the proposed framework, as they enable the estimation of tangent spaces, metric structure, and higher-order geometric quantities directly from sampled data, without requiring an explicit parametrization of the underlying manifold.

\noindent
In the proposed framework, we approximate the local metric structure at a point $\mathbf{x}_i$ using the inverse of the covariance matrix, $\boldsymbol{\Sigma}_i^{-1}$, estimated from its corresponding neighborhood patch $P_i$. This choice is well grounded in both statistical learning and differential geometry, as the inverse covariance, also known as the \emph{precision matrix}, naturally defines a local, data-adaptive inner product that captures the anisotropic structure of the data distribution \citep{MetricLearning1,MetricLearning2,MultivariateBook}. 

First, since the covariance matrix $\boldsymbol{\Sigma}_i$ is symmetric positive-definite under mild conditions (e.g., non-degenerate local neighborhoods), its inverse $\boldsymbol{\Sigma}_i^{-1}$ also defines a valid positive-definite bilinear form, thereby satisfying the fundamental requirement of a metric tensor. This ensures that local distances and angles are well-defined within each neighborhood. Second, the precision matrix encodes fine-grained information about feature dependencies: directions associated with low variance (i.e., more constrained directions) are amplified, while directions of high variance are attenuated. As a result, $\boldsymbol{\Sigma}_i^{-1}$ provides a more informative notion of distance than standard Euclidean metrics, reflecting the underlying statistical structure of the data.

Furthermore, this construction naturally captures non-isotropic behavior, which is essential for modeling data lying on curved or anisotropic manifolds. In contrast to isotropic metrics, the inverse covariance adapts to local variations in scale and orientation, effectively acting as a directional weighting mechanism in the tangent space. This perspective is closely related to the Mahalanobis distance, defined as $d_M(\mathbf{x}, \mathbf{y}) = \sqrt{(\mathbf{x}-\mathbf{y})^\top \boldsymbol{\Sigma}_i^{-1} (\mathbf{x}-\mathbf{y})}$, which generalizes Euclidean distance by incorporating feature correlations and local geometry.

From a geometric standpoint, this approximation establishes a direct connection between statistical estimation and Riemannian geometry: the inverse covariance matrix serves as a discrete analogue of the metric tensor, enabling the measurement of intrinsic quantities on the data manifold. This formulation is particularly advantageous in high-dimensional settings, where explicit manifold parametrizations are unavailable, and aligns with recent advances in metric learning that exploit local covariance structure to define adaptive geometries in data-driven spaces.

\noindent
We begin by estimating the local metric structure at each sample $\mathbf{x}_i$ through the empirical covariance matrix computed over its neighborhood patch $P_i$. Specifically, we define
\begin{equation}
	\boldsymbol{\Sigma}_i = \frac{1}{k} \sum_{\mathbf{x}_j \in P_i} (\mathbf{x}_j - \mathbf{x}_i)(\mathbf{x}_j - \mathbf{x}_i)^\top,
\end{equation}
which captures the local dispersion and directional variability of the data around $\mathbf{x}_i$. Under mild regularity conditions, $\boldsymbol{\Sigma}_i$ is positive-definite, and its inverse $\boldsymbol{\Sigma}_i^{-1}$ provides a discrete approximation of the first fundamental form (metric tensor), i.e., $\mathbb{I}_i \approx \boldsymbol{\Sigma}_i^{-1}$. This construction induces a local Mahalanobis geometry that adapts to anisotropies in the data distribution.

To approximate the second fundamental form, we draw on the connection between curvature and second-order derivatives. In particular, for manifolds represented locally as graphs, the Hessian operator encodes curvature information and is closely related to the coefficients of the second fundamental form \citep{Second_Hessian,chase2012fundamental}. Following the strategy of Hessian-based manifold learning methods such as Hessian Eigenmaps \citep{HessianEig}, we estimate a discrete local Hessian operator from each patch $P_i$.

Let $\{\mathbf{u}_1, \ldots, \mathbf{u}_m\}$ denote the eigenvectors of $\boldsymbol{\Sigma}_i$, which approximate a local orthonormal basis of the tangent space. We construct a local design matrix $X_i$ whose columns encode first- and second-order polynomial terms in this coordinate system. For the general case, $X_i$ has $1 + m + \frac{m(m+1)}{2}$ columns, corresponding to constant, linear, and quadratic terms, respectively. For example, when $m=3$, we have
\begin{equation}
	X_i = \left[ \mathbf{1}, \mathbf{u}_1, \mathbf{u}_2, \mathbf{u}_3, \mathbf{u}_1^2, \mathbf{u}_2^2, \mathbf{u}_3^2, \mathbf{u}_1 \odot \mathbf{u}_2, \mathbf{u}_1 \odot \mathbf{u}_3, \mathbf{u}_2 \odot \mathbf{u}_3 \right],
\end{equation}
where $\odot$ denotes the element-wise (Hadamard) product. This construction enables the estimation of second-order variations within the local neighborhood.

From $X_i$, we extract the components associated with second-order terms to build a local Hessian estimator. Let $H_i$ denote the matrix formed by the corresponding columns of $X_i$. We then define the local Hessian-based approximation of the second fundamental form as
\begin{equation}
	\mathcal{H}_i = H_i H_i^\top,
\end{equation}
which yields a symmetric positive semi-definite matrix encoding second-order geometric information. Accordingly, we approximate $\mathbb{II}_i \approx \mathcal{H}_i$.

Combining first- and second-order estimates, we obtain a discrete approximation of the shape operator:
\begin{equation}
	\mathcal{S}_i = \mathbb{II}_i \, \mathbb{I}_i^{-1} \;\approx\; \mathcal{H}_i \boldsymbol{\Sigma}_i,
\end{equation}
which provides a local linear operator capturing curvature effects. The mean curvature at $\mathbf{x}_i$ is then defined as the trace of $\mathcal{S}_i$:
\begin{equation}
	K_i = \mathrm{tr}(\mathcal{S}_i).
\end{equation}

This formulation admits a useful interpretation. Since $\mathcal{S}_i$ is the product of a second-order term ($\mathcal{H}_i$) and a first-order term ($\boldsymbol{\Sigma}_i$), large values of $K_i$ arise when both local dispersion and second-order variation are significant. In practice, this corresponds to regions where the data exhibits low density combined with strong geometric deformation (e.g., concave or convex transitions), which are characteristic of boundary points. Conversely, samples located in dense, geometrically smooth regions tend to exhibit low curvature values. After computing $\{K_i\}_{i=1}^n$, we normalize the curvature scores to the interval $[0,1]$ to enable consistent interpretation across datasets.

Figure~\ref{fig:outliers} provides an intuitive illustration of the interplay between local density and geometric deformation in the proposed curvature-based framework. In particular, it highlights how regions with low sampling density and pronounced concavity or convexity tend to produce higher curvature estimates, whereas densely sampled and locally flat regions yield lower curvature values. This distinction is central to our approach, as it enables the identification of boundary points as those simultaneously exhibiting geometric irregularity and reduced local support.


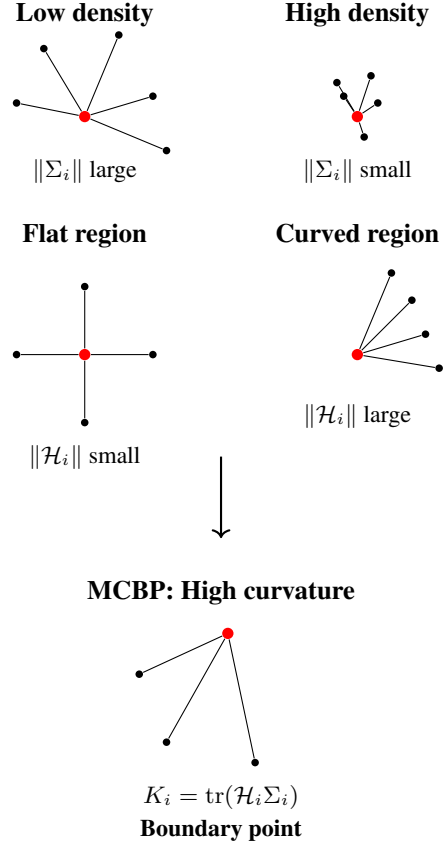
\begin{figure}
	\centering
	\begin{tikzpicture}[scale=0.9, every node/.style={font=\small}]
		
		\tikzstyle{centerpoint} = [circle, fill=red, inner sep=1.5pt]
		\tikzstyle{neighbor} = [circle, fill=black, inner sep=1pt]
		\tikzstyle{edge} = [-, thin]
		\tikzstyle{label} = [font=\small]
		\tikzstyle{title} = [font=\bfseries]
		
		\node[title] at (0,4) {Low density};
		\node[centerpoint] (c1) at (0,2.5) {};
		\foreach \x/\y in {1/0.3, -1/0.2, 0.5/1.2, -0.6/1.0, 1.2/-0.5}
		{
			\node[neighbor] (p) at (\x,2.5+\y) {};
			\draw[edge] (c1) -- (p);
		}
		\node[label] at (0,1.7) {$\|\Sigma_i\|$ large};
		
		\node[title] at (4,4) {High density};
		\node[centerpoint] (c2) at (4,2.5) {};
		\foreach \x/\y in {0.3/0.2, -0.2/0.3, 0.2/0.6, -0.3/0.5, 0.1/-0.3}
		{
			\node[neighbor] (p) at (4+\x,2.5+\y) {};
			\draw[edge] (c2) -- (p);
		}
		\node[label] at (4,1.7) {$\|\Sigma_i\|$ small};
		
		\node[title] at (0,0.75) {Flat region};
		\node[centerpoint] (c3) at (0,-1) {};
		\foreach \x/\y in {1/0, -1/0, 0/1, 0/-1}
		{
			\node[neighbor] (p) at (\x,-1+\y) {};
			\draw[edge] (c3) -- (p);
		}
		\node[label] at (0,-2.5) {$\|\mathcal{H}_i\|$ small};
		
		\node[title] at (4,0.75) {Curved region};
		\node[centerpoint] (c4) at (4,-1) {};
		\foreach \x/\y in {0.5/1.2, 0.8/0.8, 1.0/0.3, 1.2/-0.2}
		{
			\node[neighbor] (p) at (4+\x,-1+\y) {};
			\draw[edge] (c4) -- (p);
		}
		\node[label] at (4,-1.9) {$\|\mathcal{H}_i\|$ large};
		
		\draw[->, thick] (2,-2.5) -- (2,-3.7);
		
		\node[title] at (2,-4.5) {MCBP: High curvature};
		\node[centerpoint] (cf) at (2.1,-5.1) {};
		\foreach \x/\y in {-1.2/0.1, 0.5/-1.2, -0.8/-0.9}
		{
			\node[neighbor] (p) at (2+\x,-5.8+\y) {};
			\draw[edge] (cf) -- (p);
		}
		
		\node[label] at (2,-7.5) {$K_i = \mathrm{tr}(\mathcal{H}_i \Sigma_i)$};
		\node[label] at (2,-8.0) {\textbf{Boundary point}};
		
	\end{tikzpicture}
	
	\caption{
		Minimal illustration of the Mean Curvature Boundary Points (MCBP) principle. 
		Top: covariance captures local density. Middle: Hessian captures geometric deformation. 
		Bottom: high curvature emerges from the combination of low density and strong curvature, identifying boundary points.
	}
	\label{fig:mcbp_concept}
\end{figure}

Algorithm~\ref{alg:cap} summarizes the proposed \textit{Mean Curvature Boundary Points} (MCBP) procedure. The method operates by estimating a curvature score for each sample based on local first- and second-order geometric structure, followed by a percentile-based thresholding step to identify boundary points. Notably, the algorithm is fully data-driven and does not require explicit manifold parametrization, making it suitable for high-dimensional settings.

\begin{algorithm}
	\caption{Mean Curvature Boundary Points (MCBP)}\label{alg:cap}
	\begin{algorithmic}
		\Function{MCBP}{$X, k, p$}
		\State \textbf{Input:} $X \in \mathbb{R}^{n \times m}$ (dataset), $k$ (number of neighbors), $p \in (0,1)$ (percentile threshold)
		\State \textbf{Output:} $B \in \{0,1\}^n$ (boundary labels)
		\State $A \gets k$NN-Graph$(X, k)$ \Comment{Construct neighborhood graph}
		\For{$i = 1$ to $n$}
		\State $\mathcal{N}_i \gets$ neighbors of $\mathbf{x}_i$ in $A$
		\State $\boldsymbol{\Sigma}_i \gets \frac{1}{k} \sum_{\mathbf{x}_j \in \mathcal{N}_i} (\mathbf{x}_j - \mathbf{x}_i)(\mathbf{x}_j - \mathbf{x}_i)^\top$
		\State $U_i \gets$ eigenvectors of $\boldsymbol{\Sigma}_i$ \Comment{Local tangent basis}
		\State Construct local design matrix $X_i$ (constant, linear, quadratic terms)
		\State Extract second-order components $H_i$ from $X_i$
		\State $\mathcal{H}_i \gets H_i H_i^\top$ \Comment{Second fundamental form approximation}
		\State $\mathcal{S}_i \gets \mathcal{H}_i \boldsymbol{\Sigma}_i$ \Comment{Shape operator approximation}
		\State $K_i \gets \mathrm{tr}(\mathcal{S}_i)$ \Comment{Mean curvature estimate}
		\EndFor
		\State $K \gets (K - \min K)/(\max K - \min K)$ \Comment{Normalization}
		\State $T \gets$ percentile$(K, p)$
		\For{$i = 1$ to $n$}
		\If{$K_i \geq T$}
		\State $B_i \gets 1$ \Comment{Boundary point}
		\Else
		\State $B_i \gets 0$ \Comment{Interior point}
		\EndIf
		\EndFor
		\State \Return $B$
		\EndFunction
	\end{algorithmic}
\end{algorithm}


The proposed procedure differs fundamentally from classical boundary detection methods: instead of relying solely on density variations, it integrates first- and second-order geometric information into a unified curvature-based criterion. This enables a more expressive characterization of boundary regions, particularly in high-dimensional and non-linear settings where density-based assumptions are insufficient.

\subsection{Computational Complexity}
\noindent
The computational cost of the proposed MCBP algorithm is primarily driven by the construction of local neighborhoods and the estimation of curvature via local linear algebra operations. Let $n$ denote the number of samples and $m$ the dimensionality of the feature space.

The first stage consists of building the $k$-nearest neighbor graph. Using space-partitioning data structures such as KD-trees, this step can be performed in $O(n m \log n)$ time under standard assumptions \cite{scikit-learn}. In high-dimensional regimes where KD-trees degrade, approximate nearest neighbor methods may be employed, often yielding sublinear query time in practice.

The main computational burden arises from the per-sample processing loop, which is executed $n$ times. For each sample $\mathbf{x}_i$, the algorithm performs: (i) neighborhood retrieval, $O(mk)$ assuming precomputed neighbors; (ii) covariance estimation, $O(m^2 k)$; (iii) eigendecomposition of the $m \times m$ covariance matrix, $O(m^3)$; and (iv) matrix multiplications to construct the Hessian approximation and shape operator, also $O(m^3)$. The computation of the trace is negligible, requiring $O(m)$ operations. Therefore, the per-sample cost is dominated by $O(m^3)$, yielding an overall complexity of
\begin{equation}
	T(n) = O(n m \log n) + O(n m^3).
\end{equation}

It is important to note that the dependence on $k$ is linear and typically modest (e.g., $k \ll n$), and thus does not affect the asymptotic behavior. Moreover, all per-sample computations are independent and can be parallelized, which significantly improves scalability in practice.

From a practical standpoint, the algorithm scales more favorably with respect to the number of samples than to the ambient dimensionality. In high-dimensional settings, it is often beneficial to incorporate a preliminary dimensionality reduction step, such as Principal Component Analysis (PCA), to project the data onto a lower-dimensional subspace of intrinsic dimension $d \ll m$. In this case, the dominant term becomes $O(n d^3)$, which can substantially reduce computational cost while preserving the relevant geometric structure.

Overall, despite involving second-order geometric estimations, the MCBP algorithm exhibits polynomial complexity and remains computationally tractable for moderately large datasets, particularly when combined with efficient nearest neighbor search and dimensionality reduction techniques.

\subsection{Mean Curvature-Based Dataset Filtering}

\noindent
An immediate and practically relevant consequence of the proposed curvature-based framework is a principled data filtering mechanism, which we argue constitutes one of the central contributions of this work. In contrast to traditional preprocessing techniques that rely solely on density or distance criteria, our approach leverages intrinsic geometric information to selectively remove samples that lie in regions of high curvature, i.e., points associated with sharp transitions, boundary effects, or structural irregularities in the data manifold.

The underlying intuition is analogous to low-pass filtering in signal and image processing: boundary points, characterized by high mean curvature, can be interpreted as high-frequency components of the data distribution, whereas interior points correspond to low-frequency, smooth regions. By removing high-curvature samples, we obtain a geometrically “smoothed” representation of the dataset that preserves the core structure of clusters while attenuating noise, outliers, and boundary ambiguities. This perspective provides a novel bridge between differential geometry and data preprocessing, extending classical filtering concepts to high-dimensional, non-Euclidean domains. Formally, given a dataset $X \in \mathbb{R}^{n \times m}$, the proposed \emph{Mean Curvature Dataset Filtering} procedure is defined as follows:

\begin{enumerate}
	\item Construct a $k$-nearest neighbor graph over $X$ using an appropriate distance metric (e.g., Euclidean distance);
	\item Estimate the mean curvature $K_i$ for each sample $\mathbf{x}_i$ using the MCBP algorithm;
	\item Partition the dataset into two disjoint subsets based on a curvature threshold:
	\begin{itemize}
		\item $S = \{ \mathbf{x}_i \in X \;:\; K_i < T \}$, the \emph{smoothed set}, containing $(n - n_B)$ low-curvature samples;
		\item $B = \{ \mathbf{x}_i \in X \;:\; K_i \geq T \}$, the \emph{boundary set}, containing $n_B$ high-curvature samples.
	\end{itemize}
\end{enumerate}

This decomposition enables a range of downstream applications. In clustering, for instance, applying standard algorithms (e.g., $k$-means or spectral clustering) to the filtered set $S$ typically results in more stable centroid initialization, improved cluster separation, and reduced sensitivity to noise. The boundary set $B$, rather than being discarded, may also be exploited for complementary tasks such as anomaly detection, boundary refinement, or semi-supervised labeling.

Importantly, this filtering strategy is entirely data-driven and adapts to the intrinsic geometry of the dataset, without requiring strong distributional assumptions. As such, it provides a robust and flexible preprocessing tool that enhances the performance of a wide range of learning algorithms, particularly in scenarios involving complex, high-dimensional, or non-linearly structured data.

\subsection{Theoretical Justification for Curvature as a Boundary Surrogate}
\label{sec:theory}

Having introduced the discrete estimator of mean curvature employed by MCBP, we now provide
a theoretical justification for its use as a surrogate for boundary characterization. The
central premise is that regions associated with cluster transitions, low-density interfaces,
or geometric irregularities exhibit elevated curvature when the data is viewed as samples
from an underlying manifold.

\paragraph{Curvature as a geometric indicator of transition regions.}
Let $\mathcal{M} \subset \mathbb{R}^m$ be a compact, $C^3$-smooth, $d$-dimensional Riemannian
submanifold (possibly with boundary $\partial \mathcal{M}$), and let $p \in \mathcal{M}$.
The mean curvature $H(p)$ is defined as:
\begin{equation}
	H(p) = \frac{1}{d} \sum_{i=1}^{d} \kappa_i(p),
\end{equation}
where $\kappa_i(p)$ are the principal curvatures. A classical result in geometric measure theory relates mean curvature to the first variation of volume (see, e.g., \cite{Simon1983, do_carmo_differential_2016}):
\begin{equation}
	\delta_X \mathcal{H}^d(\mathcal{M}) = -\int_{\mathcal{M}} H(p)\,\langle X(p), \mathbf{n}(p)\rangle \, d\mathcal{H}^d(p).
\end{equation}
This identity shows that $H(p)$ governs the sensitivity of the manifold to infinitesimal
normal perturbations. Regions with large curvature correspond to locations where small
deformations induce significant geometric variation.

In data analysis, samples are often drawn from mixtures supported on subsets
$\mathcal{M}_1, \ldots, \mathcal{M}_C$. The union $\mathcal{M} = \bigcup_{c=1}^C \mathcal{M}_c$ may fail to be smooth at interfaces. At such locations, curvature is not classically defined, but empirical geometric estimators exhibit increased variability. This phenomenon is well documented in manifold learning and geometric inference \cite{Aamari2019, Genovese2012}, supporting the use of curvature as a proxy for boundary regions.

\paragraph{Consistency of the discrete curvature estimator.}
Let $\{x_1, \ldots, x_n\}$ be i.i.d.\ samples from a distribution supported on $\mathcal{M}$
with density $f$ bounded away from zero and infinity. Define the local covariance:
\begin{equation}
	\boldsymbol{\Sigma}_i = \frac{1}{k} \sum_{x_j \in \eta_i} (x_j - x_i)(x_j - x_i)^\top,
\end{equation}
where $\eta_i$ denotes the $k$-nearest neighbors of $x_i$. Under standard assumptions ($k \to \infty$, $k/n \to 0$), it is known that local covariance matrices converge to the tangent space structure (see \cite{Singer2012, AamariLevrard2019}):
\begin{equation}
	\boldsymbol{\Sigma}_i = \mathcal{O}(\varepsilon^2)\,\mathbf{G}(x_i) + o(\varepsilon^2),
\end{equation}
where $\varepsilon$ is the neighborhood radius and $\mathbf{G}$ is the metric tensor. The second-order estimator used in MCBP is obtained via local polynomial fitting, which has
been shown to recover second-order geometric information under suitable conditions
\cite{Aamari2019, aizenbud2021nonparametric}. Denoting this estimator by $\mathbf{H}_i$, we define:
\begin{equation}
	\widehat{K}_i = \mathrm{tr}(\mathbf{H}_i \boldsymbol{\Sigma}_i).
\end{equation}

While $\mathbf{H}_i$ does not explicitly reconstruct the second fundamental form, it
provides a consistent proxy for local curvature. Under smoothness and sampling assumptions,
it follows that:
\begin{equation}
	\widehat{K}_i = c\, H(x_i) + o_{\mathbb{P}}(1),
\end{equation}
for some constant $c > 0$, consistent with results on curvature estimation from point clouds
\cite{AamariLevrard2019}.

\paragraph{Interaction between geometry and density.}
The estimator $\widehat{K}_i$ implicitly combines geometric and statistical information.
For $k$-NN neighborhoods, the radius satisfies \cite{Devroye1996}:
\begin{equation}
	\varepsilon(x_i) \asymp \left(\frac{k}{n f(x_i)}\right)^{1/d},
\end{equation}
implying:
\begin{equation}
	\|\boldsymbol{\Sigma}_i\|_F \propto f(x_i)^{-2/d}.
\end{equation}

Thus, low-density regions produce larger covariance magnitudes, while $\mathbf{H}_i$
captures geometric bending. The curvature score admits the heuristic interpretation:
\begin{equation}
	\widehat{K}_i \;\approx\; \|\mathbf{H}_i\| \cdot \|\boldsymbol{\Sigma}_i\|,
\end{equation}
indicating that large values arise from curvature, sparsity, or both. This aligns with
observations in geometric data analysis where boundaries often coincide with regions of
both low density and high geometric variation \cite{Genovese2012}.

%
%
\smallskip
Taken together, these results support the use of curvature as a principled surrogate for
boundary detection: (i) curvature characterizes geometric sensitivity via variational
principles; (ii) the proposed estimator is consistent under standard sampling assumptions;
and (iii) it integrates geometric and density information, providing a richer signal than
classical approaches.

\subsection{The Laplace-Beltrami Operator and its Connection to Curvature}
\label{sec:laplace_beltrami}

The Laplace--Beltrami operator constitutes one of the central objects of analysis on
Riemannian manifolds, providing a canonical generalization of the classical Euclidean
Laplacian to curved spaces. Its spectral properties encode fundamental geometric and
topological information about the manifold, and its discrete counterparts, graph
Laplacians constructed from data, form the backbone of a wide range of manifold
learning and geometric deep learning methods \cite{bronstein2017geometric, coifman2006diffusion}. In the
context of the present work, this operator plays a twofold role: it provides an
independent, spectrally grounded justification for the use of mean curvature as a boundary
descriptor (via the identity established in Section~\ref{sec:theory}), and it connects the
proposed MCBP framework to the broader landscape of graph-based learning algorithms.

\paragraph{Definition and intrinsic formulation.}
Let $(\mathcal{M}, g)$ be a smooth, oriented, $d$-dimensional Riemannian manifold embedded
in $\mathbb{R}^m$, where $g$ denotes the Riemannian metric induced by the ambient space.
The \emph{Laplace--Beltrami operator} $\Delta_{\mathcal{M}}$ is the intrinsic, self-adjoint,
second-order differential operator defined as the composition of the divergence and gradient
operators on $\mathcal{M}$:
\begin{equation}
	\Delta_{\mathcal{M}} = \mathrm{div}_{\mathcal{M}} \circ \nabla_{\mathcal{M}}.
	\label{eq:lb_def}
\end{equation}
More explicitly, given a smooth function $f: \mathcal{M} \to \mathbb{R}$ and a local
coordinate chart $(u^1, \ldots, u^d)$ with metric tensor $\mathbf{G} = [g_{ij}]$, the
Laplace--Beltrami operator takes the form:
\begin{equation}
	\Delta_{\mathcal{M}} f
	= \frac{1}{\sqrt{\det \mathbf{G}}}
	\sum_{i,j=1}^{d}
	\frac{\partial}{\partial u^i}
	\left(
	\sqrt{\det \mathbf{G}}\; g^{ij}\, \frac{\partial f}{\partial u^j}
	\right),
	\label{eq:lb_local}
\end{equation}
where $g^{ij}$ denotes the $(i,j)$-entry of $\mathbf{G}^{-1}$. When $\mathcal{M} =
\mathbb{R}^d$ (i.e., $g_{ij} = \delta_{ij}$), the expression in
Eq.~\eqref{eq:lb_local} reduces to the standard Laplacian $\Delta f = \sum_{i=1}^{d}
\partial^2 f / \partial u_i^2$, confirming that $\Delta_{\mathcal{M}}$ is its natural
geometric generalization. For a $d$-dimensional submanifold of $\mathbb{R}^m$, the
operator is intrinsic: it depends only on the metric $g$ and not on the particular
embedding, making it a fundamental tool for geometry-aware data analysis.

A key property of $\Delta_{\mathcal{M}}$ is its \emph{self-adjointness} with respect to
the $L^2(\mathcal{M}, d\mathrm{vol}_g)$ inner product. That is, for any two smooth
compactly supported functions $f, h: \mathcal{M} \to \mathbb{R}$:
\begin{equation}
	\int_{\mathcal{M}} f\, \Delta_{\mathcal{M}} h\; d\mathrm{vol}_g
	= \int_{\mathcal{M}} h\, \Delta_{\mathcal{M}} f\; d\mathrm{vol}_g
	= -\int_{\mathcal{M}} \langle \nabla_{\mathcal{M}} f,\,
	\nabla_{\mathcal{M}} h \rangle_g\; d\mathrm{vol}_g,
	\label{eq:lb_selfadjoint}
\end{equation}
where the second equality follows from integration by parts on $\mathcal{M}$ (Green's
identity). This identity implies that $-\Delta_{\mathcal{M}}$ is a positive semi-definite
operator, whose spectrum $0 = \lambda_0 \leq \lambda_1 \leq \lambda_2 \leq \cdots$
consists of non-negative eigenvalues with finite multiplicity, associated with an
orthonormal basis of eigenfunctions $\{\phi_k\}$ satisfying:
\begin{equation}
	-\Delta_{\mathcal{M}} \phi_k = \lambda_k \phi_k.
	\label{eq:lb_eigen}
\end{equation}
The eigenvalues $\{\lambda_k\}$ encode rich geometric information: the multiplicity of
$\lambda_0 = 0$ equals the number of connected components of $\mathcal{M}$, the
\emph{spectral gap} $\lambda_1$ controls the rate of diffusion and the degree of cluster
separation \cite{Chung1997}, and the full spectrum determines the manifold up to isometry
in many practically relevant settings (Kac's question \cite{Kac1966}).

\paragraph{The embedding identity and mean curvature.}
A particularly illuminating consequence of the Laplace--Beltrami operator, which directly
motivates its role in the MCBP framework, is its action on the coordinate functions of the
embedding. Let $\iota = (\iota_1, \ldots, \iota_m): \mathcal{M} \hookrightarrow
\mathbb{R}^m$ denote the isometric embedding map. Applying $\Delta_{\mathcal{M}}$
componentwise to $\iota$ yields the classical identity \cite{do_carmo_differential_2016}:
\begin{equation}
	\Delta_{\mathcal{M}}\, \iota = d\, H\, \mathbf{n},
	\label{eq:lb_mean_curv}
\end{equation}
where $H: \mathcal{M} \to \mathbb{R}$ is the mean curvature function,
$\mathbf{n}: \mathcal{M} \to \mathbb{R}^m$ is the unit outward normal field, and $d$ is
the intrinsic dimension. Identity~\eqref{eq:lb_mean_curv} reveals a profound connection
between analysis and geometry: the Laplace--Beltrami operator detects mean curvature
directly from the coordinate representation of the embedding, without requiring explicit
computation of principal curvatures or the shape operator. In particular, $H(p) = 0$ if
and only if $\Delta_{\mathcal{M}} \iota(p) = \mathbf{0}$, i.e., $p$ is a \emph{minimal
	point} where the embedding is harmonic. Conversely, large values of $|\Delta_{\mathcal{M}}
\iota(p)|$ correspond to strong local bending, providing a spectral characterization of
high-curvature regions that is entirely consistent with the boundary detection criterion
employed by MCBP.

\paragraph{Discretization: from the manifold to the graph Laplacian.}
In practice, the underlying manifold $\mathcal{M}$ is not known explicitly; only a finite
set of samples $X = \{x_1, \ldots, x_n\} \subset \mathbb{R}^m$ drawn from a distribution
supported on $\mathcal{M}$ is available. It is therefore essential to construct a discrete
analogue of $\Delta_{\mathcal{M}}$ directly from the data. The standard approach consists
of building a weighted graph $\mathcal{G} = (V, E, W)$ over $X$, where $V = \{1, \ldots,
n\}$, and defining edge weights via a kernel function. Two widely used constructions are
the \emph{unnormalized} and \emph{normalized graph Laplacians} \cite{vonLuxburg2007}.

Given a symmetric weight matrix $\mathbf{W} \in \mathbb{R}^{n \times n}$ with entries
$w_{ij} \geq 0$ (e.g., $w_{ij} = \exp(-\|x_i - x_j\|^2 / 2\sigma^2)$ for a Gaussian
kernel with bandwidth $\sigma > 0$, restricted to $k$-nearest neighbor pairs), define the
diagonal degree matrix $\mathbf{D} = \mathrm{diag}(d_1, \ldots, d_n)$ with $d_i =
\sum_j w_{ij}$. The three standard graph Laplacian matrices are then:
\begin{align}
	\mathbf{L} &= \mathbf{D} - \mathbf{W}
	&& \text{(unnormalized)}, \label{eq:L_unnorm} \\
	\mathbf{L}_{\mathrm{sym}} &= \mathbf{I} - \mathbf{D}^{-1/2}\mathbf{W}\mathbf{D}^{-1/2}
	&& \text{(symmetric normalized)}, \label{eq:L_sym} \\
	\mathbf{L}_{\mathrm{rw}} &= \mathbf{I} - \mathbf{D}^{-1}\mathbf{W}
	&& \text{(random walk normalized)}. \label{eq:L_rw}
\end{align}
All three matrices are positive semi-definite, and their smallest eigenvalue equals zero
with eigenvector $\mathbf{1}$ (the constant vector), mirroring the continuous case. Their
non-trivial spectra approximate the spectrum of $-\Delta_{\mathcal{M}}$ as $n \to \infty$,
under appropriate conditions on the kernel bandwidth and the data distribution.

The connection between graph Laplacians and $\Delta_{\mathcal{M}}$ is made rigorous by the
following convergence result, which constitutes one of the foundational theorems of
manifold learning \cite{coifman2006diffusion, Belkin2007, Singer2006}:

\begin{theorem}[Pointwise convergence of the graph Laplacian]
	\label{thm:laplacian_convergence}
	Let $\mathcal{M}$ be a compact, smooth $d$-dimensional submanifold of $\mathbb{R}^m$ and
	let $f: \mathcal{M} \to \mathbb{R}$ be a $C^2$ function. Consider the random walk graph
	Laplacian $\mathbf{L}_{\mathrm{rw}}$ constructed from $n$ i.i.d.\ samples from a density
	$\mu$ on $\mathcal{M}$, using a Gaussian kernel with bandwidth $\sigma = \sigma(n)$.
	Suppose $\sigma \to 0$ and $n\sigma^{d+2} / \log n \to \infty$ as $n \to \infty$. Then,
	for each sample $x_i$:
	\begin{equation}
		\frac{1}{\sigma^2}
		\left[\mathbf{L}_{\mathrm{rw}} \mathbf{f}\right]_i
		\xrightarrow{\;\mathbb{P}\;}
		-\Delta_{\mathcal{M}} f(x_i) + \nabla_{\mathcal{M}} \log \mu(x_i)
		\cdot \nabla_{\mathcal{M}} f(x_i),
		\label{eq:graph_lap_convergence}
	\end{equation}
	where $\mathbf{f} = (f(x_1), \ldots, f(x_n))^\top$. For the density-normalized (diffusion
	maps) version of the operator, the density-dependent drift term vanishes, recovering a
	consistent estimator of $-\Delta_{\mathcal{M}} f$ \cite{coifman2006diffusion}.
\end{theorem}

\noindent
Theorem~\ref{thm:laplacian_convergence} has a direct and important implication for the
MCBP framework. Since $\mathbf{L}_{\mathrm{rw}}$ converges to $-\Delta_{\mathcal{M}}$
pointwise, and since identity~\eqref{eq:lb_mean_curv} connects $\Delta_{\mathcal{M}}$ to
mean curvature, it follows that computing the graph Laplacian of the coordinate functions
$\iota_\ell(x_i) = (x_i)_\ell$ yields an estimator of $H(x_i) \cdot \mathbf{n}(x_i)$.
More precisely, defining the matrix $\mathbf{Z} = [\iota(x_1) | \cdots | \iota(x_n)]^\top
\in \mathbb{R}^{n \times m}$:
\begin{equation}
	\left[\mathbf{L}_{\mathrm{rw}}\, \mathbf{Z}\right]_i
	\;\approx\; -d\, H(x_i)\, \mathbf{n}(x_i),
	\label{eq:graph_lap_curv}
\end{equation}
so that the \emph{norm of the graph Laplacian of the embedding coordinates} provides a
direct, data-driven estimate of the magnitude of mean curvature at each sample point.
The MCBP estimator $\widehat{K}_i = \mathrm{tr}(\mathbf{H}_i \boldsymbol{\Sigma}_i)$
can thus be understood as a localized, patch-based implementation of this principle: rather
than applying a global graph Laplacian, which requires solving a dense eigenvalue problem
of size $n \times n$, MCBP computes an equivalent second-order geometric quantity from
small local neighborhoods, achieving the same discriminative power at a fraction of the
computational cost.

\paragraph{Spectral interpretation and cluster structure.}
Beyond its role in curvature estimation, the Laplace--Beltrami operator provides a spectral
lens through which the cluster structure of the data can be understood. The low-frequency
eigenfunctions $\phi_1, \ldots, \phi_c$ associated with the smallest non-zero eigenvalues
$\lambda_1, \ldots, \lambda_c$ vary slowly within clusters and change abruptly at cluster
boundaries, encoding the global cluster geometry in a coordinate-free manner. This is the
spectral clustering principle \cite{vonLuxburg2007}: the leading eigenvectors of
$\mathbf{L}_{\mathrm{rw}}$ provide an embedding in which clusters become linearly separable.

Crucially, boundary points, which the MCBP algorithm identifies as high-curvature regions, correspond precisely to the samples where low-frequency eigenfunctions exhibit the
steepest gradient \cite{Nadler2006}. That is, $\|\nabla_{\mathcal{M}} \phi_k(p)\|$ is
maximized along decision boundaries. This provides a spectral complement to the
curvature-based characterization: boundary points are simultaneously the loci of (i) high
mean curvature, (ii) large $|\Delta_{\mathcal{M}} \iota|$, and (iii) steep spectral
eigenfunctions. These three characterizations are geometrically equivalent in the continuum
limit, and the fact that MCBP recovers this structure through a purely local estimator
further validates the theoretical foundations of the proposed method.

Taken together, the Laplace-Beltrami operator serves as a unifying theoretical bridge
between the differential geometric formulation of MCBP, the spectral theory of manifold
learning, and the graph-based algorithms that form the practical foundation of modern
unsupervised learning. Its connection to mean curvature via identity~\eqref{eq:lb_mean_curv},
combined with the consistency of graph Laplacian discretizations established in
Theorem~\ref{thm:laplacian_convergence}, provides a rigorous spectral justification for
the MCBP boundary detection criterion, reinforcing its status as a principled geometric
method rather than a heuristic construction.

\section{Computational experiments and results}\label{sec4}

\noindent
We conduct a comprehensive empirical evaluation to assess the effectiveness, robustness, and generality of the proposed Mean Curvature Boundary Points (MCBP) algorithm across datasets with varying geometric and statistical properties. The experimental protocol is designed to validate both the geometric intuition underlying the method and its practical utility in downstream learning tasks. To this end, we consider two complementary settings. First, we employ synthetic two-dimensional datasets with controlled shapes and boundary characteristics, which allow for direct visualization of curvature estimates and provide qualitative insight into the behavior of the method. Second, we evaluate MCBP on a diverse collection of real-world, high-dimensional datasets obtained from the OpenML repository (\url{www.openml.org}), encompassing a wide range of sample sizes, feature dimensions, and class distributions. This combination of controlled and real-world scenarios enables a thorough analysis of the proposed approach, highlighting its ability to capture meaningful geometric structure and its impact on tasks such as clustering and data preprocessing in complex, high-dimensional environments. 

\subsection{Synthetic Data}
\noindent
We begin with a controlled experiment on synthetic data to qualitatively assess the behavior of the proposed MCBP algorithm and to validate its underlying geometric intuition. Specifically, we generate a two-dimensional isotropic Gaussian dataset with $n = 400$ samples and apply MCBP using $k = \lfloor \log_2 n \rfloor$ for neighborhood construction and a percentile threshold $p = 0.8$ for boundary detection. This configuration yields a threshold value $T = 0.168$, corresponding to the 80th percentile of the empirical distribution of mean curvature estimates.

\begin{figure}
	\centering
	\includegraphics[scale=0.29]{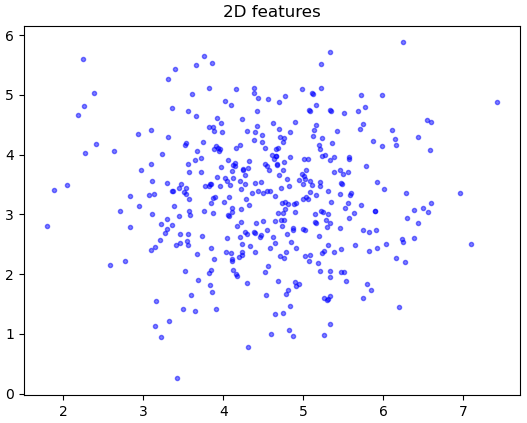}
	\includegraphics[scale=0.29]{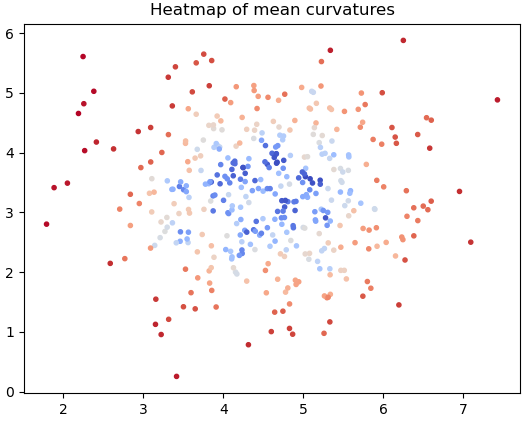}
	\includegraphics[scale=0.29]{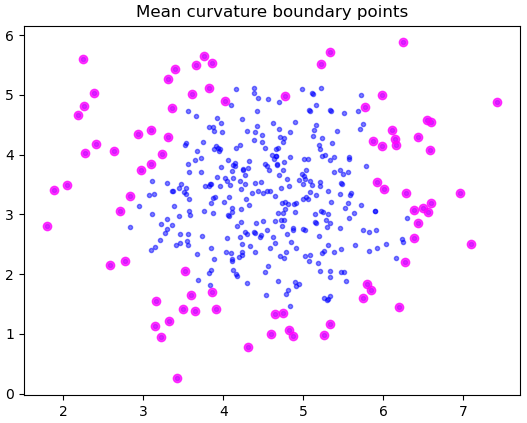}
	\caption{Results for a 2D Gaussian blob dataset. From left to right: (a) generated samples; (b) heatmap of mean curvature values; (c) boundary points identified via curvature thresholding.}
	\label{fig:blobs}
\end{figure}

Figure~\ref{fig:blobs} summarizes the results. The first panel shows the original data distribution, while the second presents a heatmap of the estimated mean curvature values, where cooler colors (blue) indicate low curvature and warmer colors (red) indicate high curvature. The third panel depicts the samples identified as boundary points after thresholding.

The results reveal a clear and consistent pattern: samples located in high-density regions near the center of the distribution exhibit low curvature values, whereas points in sparsely populated regions, particularly near the periphery, are associated with higher curvature. This behavior aligns with the theoretical interpretation of mean curvature as a joint indicator of local dispersion and geometric deformation. Importantly, it demonstrates that the proposed curvature-based criterion effectively captures boundary structure even in simple, isotropic settings, providing a strong baseline validation before proceeding to more complex and high-dimensional scenarios.

\noindent
In the second experiment, we extend the previous analysis to a multi-cluster scenario in order to evaluate the ability of the proposed MCBP algorithm to capture boundary structures between distinct data regions. To this end, we generate a dataset composed of two isotropic Gaussian blobs in $\mathbb{R}^2$ with a total of $n = 500$ samples. The MCBP algorithm is applied using $k = \lfloor \log_2 n \rfloor$ for neighborhood construction and a percentile threshold $p = 0.8$, resulting in a curvature threshold $T = 0.130$.

\begin{figure}
	\centering
	\includegraphics[scale=0.28]{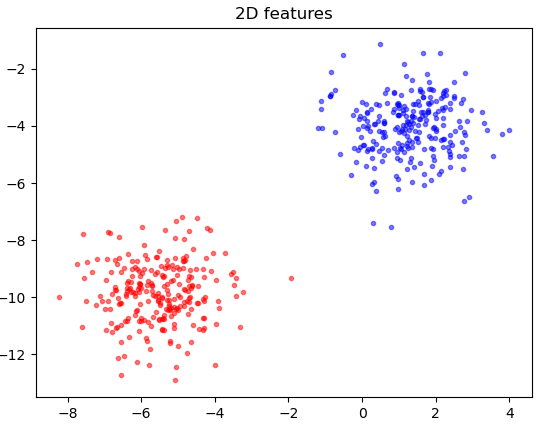}
	\includegraphics[scale=0.28]{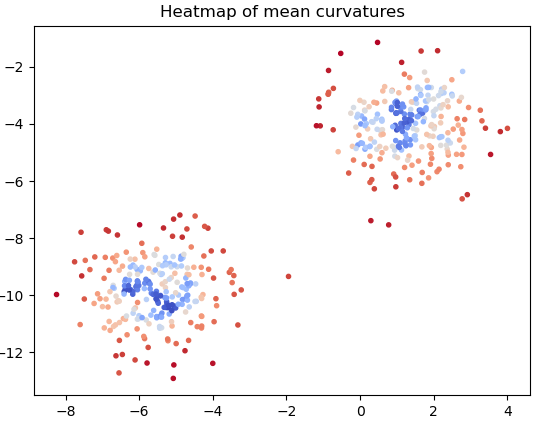}
	\includegraphics[scale=0.28]{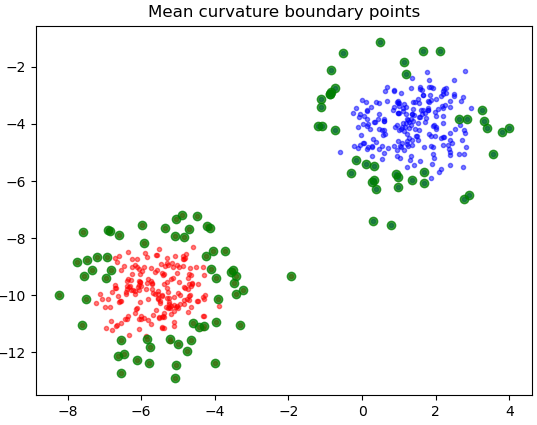}
	\caption{Results for a two-cluster 2D Gaussian dataset. From left to right: (a) generated samples; (b) heatmap of mean curvature values; (c) boundary points identified via curvature thresholding.}
	\label{fig:blobs2}
\end{figure}

Figure~\ref{fig:blobs2} presents the corresponding results. The first panel shows the generated dataset with two well-separated clusters, while the second panel illustrates the heatmap of mean curvature values across the data space. As in the previous experiment, low-curvature regions (in blue) correspond to dense, homogeneous areas within clusters, whereas high-curvature regions (in red) highlight geometrically complex or sparsely populated, particularly near cluster boundaries and outer edges. Finally, the third panel depicts the samples classified as boundary points.

Notably, the proposed method successfully identifies not only the outer contours of each cluster but also the transitional regions between them, demonstrating sensitivity to both intra-cluster geometry and inter-cluster separation. This behavior underscores the advantage of curvature-based analysis over purely density-driven approaches, as it captures subtle geometric variations that arise in multi-modal distributions. These results further support the effectiveness of MCBP as a tool for boundary detection and structure-aware preprocessing in more complex data configurations.

\noindent
In the third experiment, we consider a more challenging geometric configuration involving anisotropic structures. Specifically, we generate a dataset composed of two elliptical clusters in $\mathbb{R}^2$ with a total of $n = 500$ samples, thereby introducing directional variability and non-isotropic local geometry. The MCBP algorithm is applied using $k = \lfloor \log_2 n \rfloor$ and a percentile threshold $p = 0.75$, yielding a curvature threshold $T = 0.099$.

\begin{figure}
	\centering
	\includegraphics[scale=0.28]{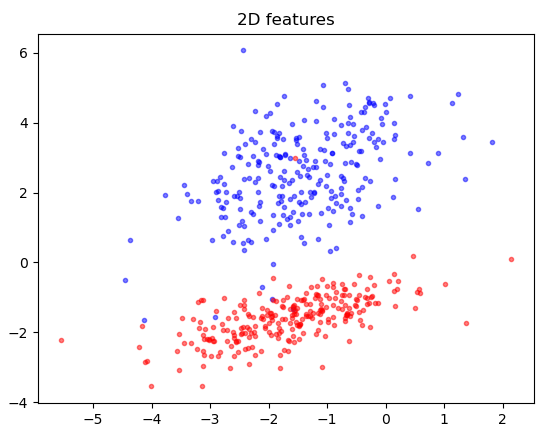}
	\includegraphics[scale=0.28]{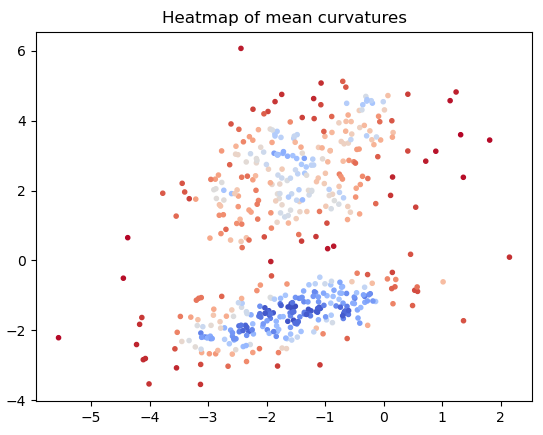}
	\includegraphics[scale=0.28]{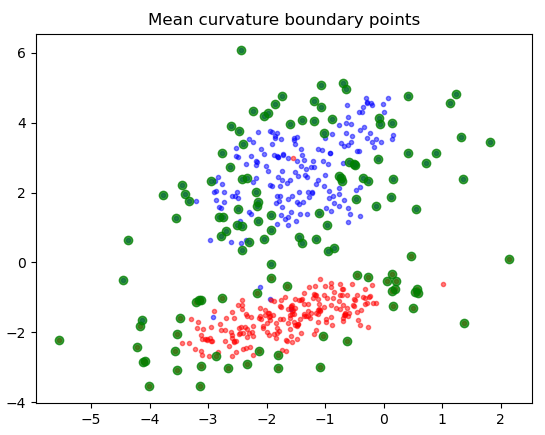}
	\caption{Results for a two-cluster anisotropic (elliptical) dataset. From left to right: (a) generated samples; (b) heatmap of mean curvature values; (c) boundary points identified via curvature thresholding.}
	\label{fig:elipses}
\end{figure}

Figure~\ref{fig:elipses} presents the results. The first panel illustrates the generated dataset, where each cluster exhibits elongated structure along a principal direction. The second panel shows the heatmap of mean curvature values, revealing a non-uniform distribution of curvature across the data manifold. In contrast to isotropic Gaussian blobs, curvature is now strongly influenced by the local anisotropy: regions along the major axes tend to exhibit lower curvature, while points near the extremities and along the minor axes display higher curvature values. Finally, the third panel highlights the boundary points obtained after thresholding.

These results emphasize an important property of the proposed method: its ability to adapt to anisotropic and direction-dependent structures. The curvature estimates reflect not only variations in local density but also changes in geometric orientation and elongation, enabling a more nuanced identification of boundary regions. In particular, MCBP successfully captures the outer contours of the elliptical clusters as well as regions of higher geometric distortion, demonstrating robustness in scenarios where classical density-based methods often struggle. This experiment further reinforces the advantage of incorporating second-order geometric information for boundary detection in non-isotropic data distributions.

\noindent
Finally, we consider a non-linearly separable scenario to evaluate the behavior of the proposed method under more complex manifold geometries. In this experiment, we generate the well-known two-moons dataset with $n = 1000$ samples, characterized by highly non-convex cluster shapes and strong curvature variations along the data manifold. The MCBP algorithm is applied with $k = \lfloor \log_2 n \rfloor$ and a percentile threshold $p = 0.75$, resulting in a threshold $T = 0.186$.

\begin{figure}
	\centering
	\includegraphics[scale=0.27]{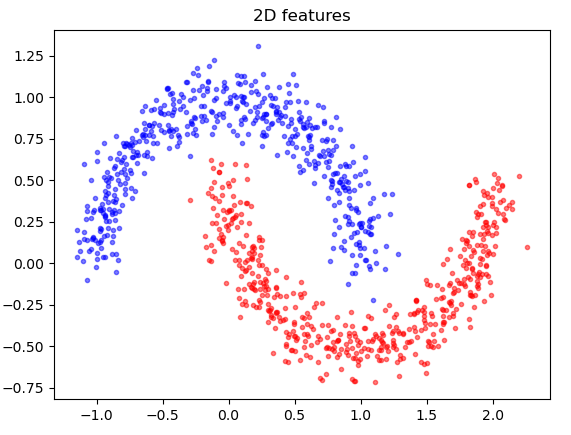}
	\includegraphics[scale=0.27]{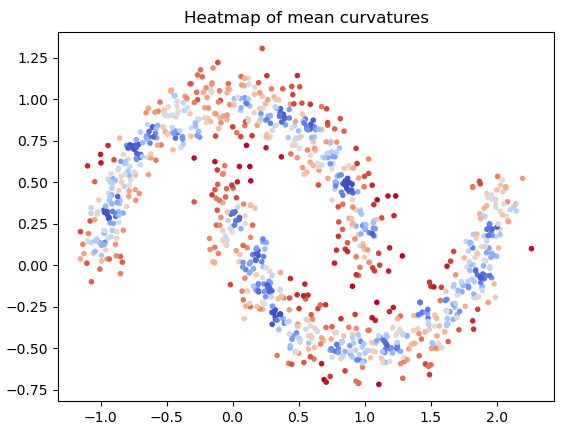}
	\includegraphics[scale=0.27]{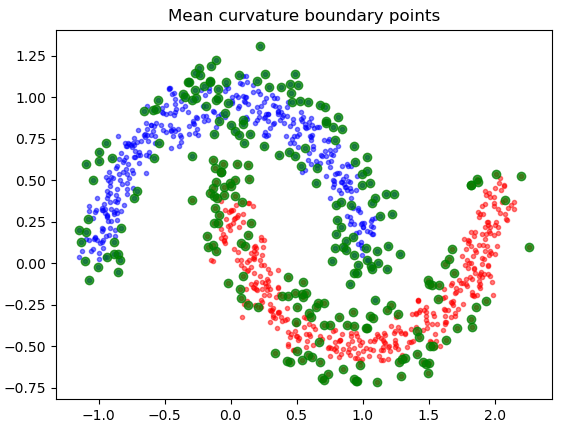}
	\caption{Results for the two-moons dataset. From left to right: (a) generated samples; (b) heatmap of mean curvature values; (c) boundary points identified via curvature thresholding.}
	\label{fig:moons}
\end{figure}

The results are depicted in Figure~\ref{fig:moons}. The first panel shows the generated dataset, where the two interleaving crescent-shaped clusters exhibit pronounced non-linearity. The second panel presents the heatmap of mean curvature values, revealing a rich geometric structure: regions with high curvature are concentrated along the outer boundaries and in areas of strong bending, while relatively flatter regions along the arcs display lower curvature values. The third panel shows the boundary points identified after thresholding.

This experiment underscores the ability of MCBP to effectively capture complex, non-linear geometric patterns that are challenging for traditional methods. In particular, the curvature-based criterion successfully identifies boundary regions along the intricate shapes of the clusters without relying on global assumptions about cluster geometry or separability. Unlike purely density-based approaches, which may struggle in regions of varying density or overlapping support, the proposed method leverages second-order geometric information to detect boundaries in a manner that is inherently sensitive to the underlying manifold structure. These results further demonstrate the robustness and versatility of MCBP in handling datasets with complex topological and geometric characteristics.

\subsection{Real High-Dimensional Datasets}

\noindent
To assess the effectiveness of the proposed MCBP algorithm in realistic and high-dimensional settings, we conducted an extensive set of experiments on real-world datasets. In contrast to synthetic scenarios, these datasets present intrinsic challenges such as high dimensionality, class overlap, noise, and complex, non-linear decision boundaries. The primary objective of this evaluation is to investigate whether the curvature-based formulation can reliably capture boundary structure in situations where traditional density-based or distance-based methods tend to degrade, particularly when class separation is ambiguous or the underlying geometry is not readily observable.

To this end, we selected 25 publicly available datasets from the OpenML repository (\url{www.openml.org}), covering a diverse range of domains, sample sizes, feature dimensions, and numbers of classes. This diversity ensures a comprehensive evaluation across different data regimes, including low-sample/high-dimensional settings, multi-class problems, and datasets with varying degrees of class imbalance and noise. All datasets were preprocessed following standard practices, including normalization to zero mean and unit variance, to ensure comparability across experiments and to stabilize the estimation of local geometric quantities.

In Table \ref{tab:datasets} we provide a brief description of each dataset, highlighting their key characteristics and relevance to the evaluation of boundary detection and curvature-based data filtering methods.

\begin{table*}
	\centering
	\small
	\caption{Summary of the datasets used in the computational experiments.}
	\begin{tabular}{lccc}
		\toprule
		\textbf{Dataset} & \textbf{\# samples} & \textbf{\# features} & \textbf{\# classes} \\
		\midrule
		AP\_Breast\_Colon         & 630              & 10935             & 2                \\
		AP\_Omentum\_Kidney       & 337              & 10935             & 2                \\
		AP\_Omentum\_Ovary        & 275              & 10935             & 2                \\
		arsenic-male-bladder      & 559              & 5                 & 14               \\
		artificial-characters     & 10218            & 7                 & 10               \\
		balance-scale             & 625              & 4                 & 3                \\
		cardiotocography          & 2126             & 35                & 10               \\
		coil-20                   & 1440             & 1024              & 20               \\
		dermatology               & 366              & 34                & 6                \\
		diabetes                  & 768              & 8                 & 2                \\
		digits                    & 1797             & 64                & 10               \\
		DLBCL                     & 77               & 5470              & 2                \\
		doggle\_table\_a2         & 310              & 8                 & 9                \\
		Engine1                   & 383              & 5                 & 3                \\
		Fashion-MNIST             & 70000            & 784               & 10               \\
		fl2000                    & 67               & 15                & 5                \\
		Flare                     & 1066             & 11                & 6                \\
		gas-drift                 & 13910            & 128               & 6                \\
		glass                     & 214              & 9                 & 6                \\
		hayes-roth                & 160              & 4                 & 3                \\
		heart-h                   & 294              & 13                & 2                \\
		Indian\_pines             & 9144             & 220               & 8                \\
		ionosphere                & 351              & 34                & 2                \\
		ipums\_la\_98-small       & 7487             & 60                & 7                \\
		ipums\_la\_99-small       & 8844             & 60                & 7                \\
		iris                      & 150              & 4                 & 3                \\
		LED-display-domain-7digit & 500              & 7                 & 10               \\
		led7                      & 3200             & 7                 & 10               \\
		letter                    & 20000            & 16                & 26               \\
		mammography               & 11183            & 6                 & 2                \\
		mfeat-factors             & 2000             & 216               & 10               \\
		mfeat-morphological       & 2000             & 6                 & 10               \\
		mfeat-pixel               & 2000             & 240               & 10               \\
		mfeat-zernike             & 2000             & 47                & 10               \\
		optdigits                 & 5620             & 64                & 10               \\
		pendigits                 & 10992            & 16                & 10               \\
		penguins                  & 344              & 6                 & 3                \\
		prnn\_synth               & 250              & 2                 & 2                \\
		prnn\_viruses             & 61               & 18                & 4                \\
		Satellite                 & 5100             & 36                & 2                \\
		satimage                  & 6430             & 36                & 6                \\
		seeds                     & 210              & 7                 & 3                \\
		segment                   & 2310             & 19                & 7                \\
		solar-flare               & 1066             & 12                & 6                \\
		steel-plates-fault        & 1941             & 33                & 2                \\
		tae                       & 151              & 5                 & 3                \\
		tecator                   & 240              & 124               & 2                \\
		texture                   & 5500             & 40                & 11               \\
		thoracic\_surgery         & 470              & 16                & 2                \\
		thyroid-dis               & 2800             & 26                & 7                \\
		tr11.wc                   & 414              & 6430              & 9                \\
		tr23.wc                   & 204              & 5833              & 6                \\
		tr41.wc                   & 878              & 7455              & 10               \\
		UMIST\_Faces\_Cropped     & 575              & 10304             & 20               \\
		USPS                      & 9298             & 256               & 10               \\
		vote                      & 435              & 17                & 2                \\
		vowel                     & 990              & 12                & 11               \\
		wine                      & 178              & 13                & 3                \\
		wine-quality-red          & 1599             & 11                & 6                \\
		\bottomrule
	\end{tabular}
	\label{tab:datasets}
\end{table*}

\noindent
Prior to curvature estimation, all datasets are standardized to zero mean and unit variance in order to ensure numerical stability and comparability across features. Additionally, for datasets with high dimensionality, we incorporate a dimensionality reduction step based on Principal Component Analysis (PCA). Specifically, when the number of features exceeds a predefined threshold (e.g., 50), the data is projected onto a lower-dimensional subspace with the first 50 principal components before applying the MCBP algorithm. This preprocessing step serves two primary purposes: (i) to reduce the computational burden associated with local covariance and Hessian estimation, whose costs scale with the ambient dimension, and (ii) to mitigate numerical instabilities arising in high-dimensional regimes, particularly in scenarios where the number of features is comparable to or exceeds the number of samples, leading to ill-conditioned covariance matrices and the well-known curse of dimensionality.

\begin{figure}
	\centering
	\includegraphics[scale=0.48]{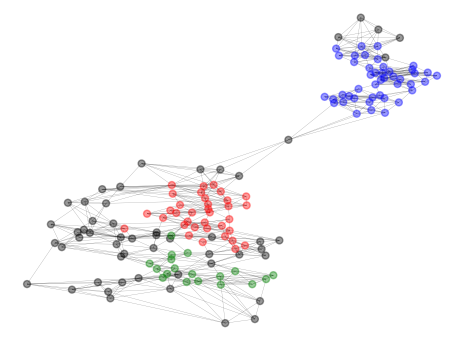}
	\includegraphics[scale=0.48]{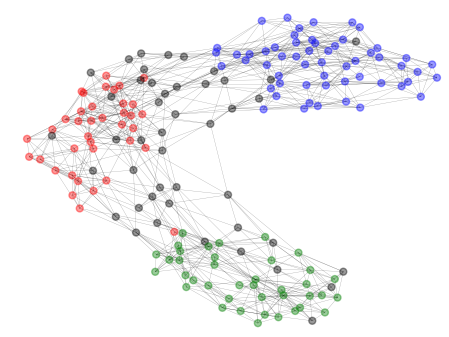}
	\includegraphics[scale=0.51]{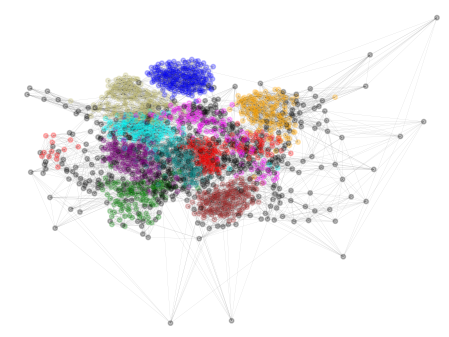}
	\includegraphics[scale=0.51]{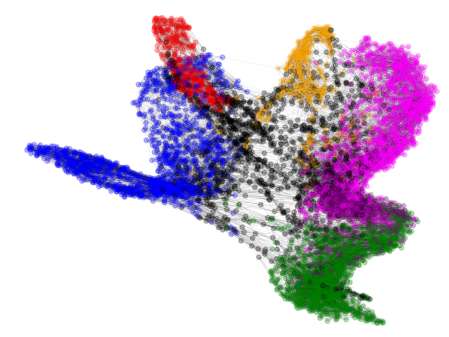}
	\includegraphics[scale=0.51]{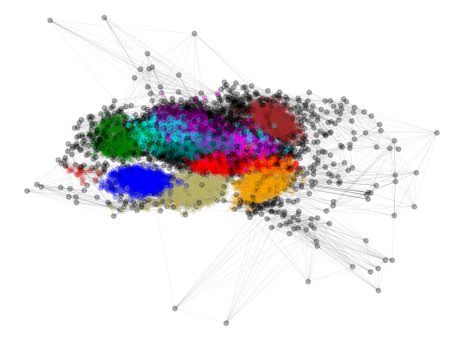}
	\includegraphics[scale=0.51]{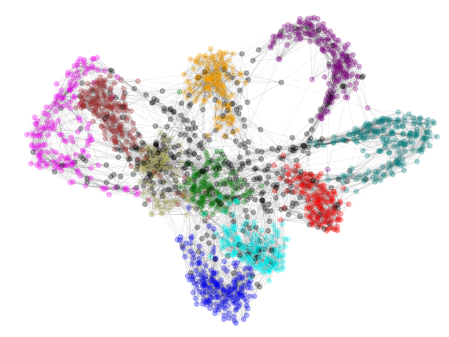}
	\caption{Boundary points detected by the proposed MCBP algorithm (black nodes) across six benchmark datasets. From top to bottom and left to right: \textit{iris}, \textit{wine}, \textit{digits}, \textit{Satellite}, \textit{optdigits}, and \textit{mfeat-pixel}. The detected points can be broadly categorized into (i) outer boundary points, often associated with outliers or low-density regions, and (ii) inner boundary points, which lie along decision boundaries between clusters.}
	\label{fig:clusters}
\end{figure}

Figure~\ref{fig:clusters} illustrates the behavior of the proposed method on six representative datasets: \textit{iris}, \textit{wine}, \textit{digits}, \textit{Satellite}, \textit{optdigits}, and \textit{mfeat-pixel}. For each dataset, we construct the corresponding $k$-nearest neighbor graph and highlight the samples identified as boundary points using a percentile threshold $p = 0.75$, i.e., selecting the top 25\% highest curvature values. The detected boundary points are shown as black nodes.

A qualitative inspection of the results reveals that MCBP consistently identifies two distinct types of boundary structures. First, in datasets characterized by the presence of outliers or sparse regions, high-curvature points tend to concentrate along the outer contours of the data distribution, effectively capturing external boundaries. Second, in datasets with well-defined cluster structure and limited noise, boundary points are predominantly located in transitional regions between clusters, delineating internal decision boundaries. This dual capability, capturing both outer and inner boundary structures, highlights a key advantage of the proposed curvature-based approach over traditional density-driven methods, which often fail to distinguish between these two regimes. As such, MCBP provides a unified and geometrically grounded mechanism for identifying structurally informative samples in both noisy and well-separated data distributions.

\noindent
In a second set of experiments, we evaluate the practical impact of the proposed mean curvature dataset filtering strategy as a preprocessing step for clustering. Specifically, for each of the 25 high-dimensional datasets, we apply the MCBP algorithm with percentile parameter $p = 0.75$, such that the threshold $T$ corresponds to the 75th percentile of the empirical distribution of mean curvature values. This results in the removal of the top 25\% highest-curvature samples, yielding a filtered dataset $S$ that emphasizes low-curvature (i.e., geometrically smoother and denser) regions of the data manifold.

Subsequently, we perform clustering using the $k$-means algorithm with $k$-means++ initialization \cite{kmeans++} on both the original dataset $X$ and its filtered counterpart $S$, using the same number of clusters $k$ in both cases. Clustering quality is assessed using three widely adopted internal validation indices: the Calinski–Harabasz index (CH) \cite{calinski}, the Davies–Bouldin index (DB) \cite{DB}, and the Silhouette Coefficient (SC) \cite{SC}. Recall that higher values of CH and SC indicate better-defined cluster structures, whereas lower values of DB correspond to improved clustering quality.

The results, reported in Table~\ref{tab:results1}, demonstrate a consistent and statistically meaningful improvement in clustering performance after the proposed filtering step. On average, the Silhouette Coefficient increases from $0.193$ to $0.288$, indicating enhanced intra-cluster cohesion and inter-cluster separation. Similarly, the Calinski–Harabasz index exhibits a substantial increase (from $693.715$ to $948.132$), suggesting improved compactness and separation of clusters in the filtered space. Conversely, the Davies–Bouldin index decreases from $2.335$ to $1.737$, reflecting a reduction in cluster overlap and dispersion.

\begin{table}
	\centering
	\caption{Internal indices obtained after k-means++ clustering of the original raw data $X$ and the filtered data $S$.}
	\begin{tabular}{ccccccc}
		\toprule
		& \multicolumn{3}{c}{\textbf{Original data}} & \multicolumn{3}{c}{\textbf{Filtered data}}            \\
		\midrule	                    
		\textbf{Datasets}           & \textbf{SC}  & \textbf{CH}  & \textbf{DB}  & \textbf{SC}    & \textbf{CH}         & \textbf{DB}    \\
		\midrule
		iris                      & 0.4799      & 157.3602            & 0.7894          & \textbf{0.5303}  & \textbf{193.6954}    & \textbf{0.6574} \\
		wine                      & 0.2840      & 70.9400             & 1.3890          & \textbf{0.3700}  & \textbf{86.8090}     & \textbf{1.0590} \\
		digits                    & 0.0171      & \textbf{28.4674}    & 3.8744          & \textbf{0.0570}  & 27.1675              & \textbf{3.5870} \\
		prnn\_viruses             & 0.2448      & 13.7203             & 1.5892          & \textbf{0.4139}  & \textbf{29.4246}     & \textbf{0.8939} \\
		diabetes                  & 0.1646      & 146.7425            & 2.0994          & \textbf{0.2324}  & \textbf{156.9266}    & \textbf{1.7507} \\
		LED-display-domain-7digit & 0.3793      & 104.6648            & 1.2722          & \textbf{0.6796}  & \textbf{252.6741}    & \textbf{0.6035} \\
		dermatology               & 0.1566      & 67.3073             & 2.2340          & \textbf{0.1883}  & \textbf{73.4740}     & \textbf{2.1060} \\
		led7                      & 0.3864      & 693.9301            & 1.2855          & \textbf{0.5801}  & \textbf{993.8856}    & \textbf{0.7647} \\
		steel-plates-fault        & 0.2767      & 515.7229            & 1.2606          & \textbf{0.3037}  & \textbf{565.4308}    & \textbf{1.1417} \\
		Flare                     & 0.2114      & 180.6994            & 1.4335          & \textbf{0.3787}  & \textbf{293.3462}    & \textbf{1.0613} \\
		pendigits                 & 0.2738      & 2,307.2487          & 1.3821          & \textbf{0.3076}  & \textbf{2,496.4991}  & \textbf{1.2895} \\
		satimage                  & 0.3522      & 4,797.1089          & 1.0234          & \textbf{0.4314}  & \textbf{6,171.9679}  & \textbf{0.7742} \\
		mfeat-pixel               & 0.0333      & \textbf{30.8179}    & 3.8825          & \textbf{0.0548}  & 29.7255              & \textbf{3.5305} \\
		mfeat-morphological       & 0.5378      & 2,836.8208          & 0.7260          & \textbf{0.5834}  & \textbf{3,458.4824}  & \textbf{0.6470} \\
		optdigits                 & 0.0140      & \textbf{90.1927}    & 3.8637          & \textbf{0.0540}  & 87.5224              & \textbf{3.4536} \\
		mammography               & 0.4904      & 5,977.7711          & 0.9609          & \textbf{0.6374}  & \textbf{11,530.2653} & \textbf{0.6572} \\
		USPS                      & -0.0549     & 147.7980            & 3.8465          & \textbf{-0.0129} & \textbf{179.0574}    & \textbf{3.0914} \\
		Satellite                 & 0.4301      & 5,216.9835          & 0.9202          & \textbf{0.5139}  & \textbf{6,185.1103}  & \textbf{0.7433} \\
		ipums\_la\_99-small       & 0.0007      & 144.1783            & 4.5469          & \textbf{0.0499}  & \textbf{169.0616}    & \textbf{3.5386} \\
		ipums\_la\_98-small       & 0.0050      & 115.3043            & 4.7772          & \textbf{0.0427}  & \textbf{146.0484}    & \textbf{3.7011} \\
		cardiotocography          & 0.1614      & 201.3592            & 1.8590          & \textbf{0.2163}  & \textbf{199.6298}    & \textbf{1.6167} \\
		gas-drift                 & 0.0860      & 201.9786            & 3.4622          & \textbf{0.1403}  & \textbf{549.3173}    & \textbf{2.1817} \\
		AP\_Breast\_Colon         & 0.0312      & 8.5064              & \textbf{2.8206} & \textbf{0.0551}  & \textbf{9.2042}      & 5.5461          \\
		vowel                     & 0.1745      & \textbf{125.5085}   & 1.6898          & \textbf{0.2163}  & 113.8239             & \textbf{1.5497} \\
		ionosphere                & 0.2704      & 95.9125             & 1.6819          & \textbf{0.3865}  & \textbf{137.0405}    & \textbf{1.2671} \\
		solar-flare               & 0.2163      & 188.5326            & 1.4525          & \textbf{0.3615}  & \textbf{273.0862}    & \textbf{1.2735} \\
		vote                      & 0.2862      & 181.3880            & 1.4480          & \textbf{0.4086}  & \textbf{277.0933}    & \textbf{1.0295} \\
		fl2000                    & 0.4121      & 40.2213             & 0.7060          & \textbf{0.5295}  & \textbf{56.2753}     & \textbf{0.5445} \\
		seeds                     & 0.4024      & 249.6550            & 0.9221          & \textbf{0.4761}  & \textbf{266.3524}    & \textbf{0.7563} \\
		coil-20                   & 0.0575      & 26.2887             & 2.6548          & \textbf{0.1799}  & \textbf{36.3722}     & \textbf{2.1835} \\
		letter                    & 0.1454      & \textbf{1,291.1190} & 1.6533          & \textbf{0.1678}  & 1,121.6117           & \textbf{1.5953} \\
		Fashion-MNIST             & 0.0448      & 1,188.3357          & 3.7282          & \textbf{0.0642}  & \textbf{1,257.6910}  & \textbf{3.2775} \\
		thyroid-dis               & 0.1513      & 206.9558            & 2.2627          & \textbf{0.2910}  & \textbf{334.5878}    & \textbf{1.0182} \\
		UMIST\_Faces\_Cropped     & 0.0932      & 10.3407             & 2.5635          & \textbf{0.1496}  & \textbf{12.3307}     & \textbf{2.1875} \\
		tr23.wc                   & 0.1073      & 4.2307              & 2.1373          & \textbf{0.5799}  & \textbf{5.9179}      & \textbf{0.2816} \\
		tr41.wc                   & 0.1361      & 18.5660             & 1.4231          & \textbf{0.1740}  & \textbf{85.9308}     & \textbf{1.1306} \\
		DLBCL                     & 0.0195      & 1.0655              & 0.9407          & \textbf{0.0878}  & \textbf{1.3696}      & \textbf{0.8255} \\
		thoracic\_surgery         & 0.1704      & \textbf{55.7060}    & 2.3208          & \textbf{0.2235}  & 54.3333              & \textbf{1.4912} \\
		AP\_Omentum\_Kidney       & 0.0513      & \textbf{4.9412}     & 6.9474          & \textbf{0.2617}  & 3.3424               & \textbf{1.4653} \\
		AP\_Omentum\_Ovary        & 0.015       & \textbf{4.198}      & 7.584           & \textbf{0.137}   & 3.391                & \textbf{3.223}  \\
		\midrule
		Average                   & 0.193       & 693.715             & 2.335           & \textbf{0.288}   & \textbf{948.132}     & \textbf{1.737}  \\
		Median                    & 0.163       & 134.843             & 1.686           & \textbf{0.247}   & \textbf{162.994}     & \textbf{1.281}  \\ 	
		\bottomrule    
		\end{tabular}
	\label{tab:results1}
\end{table}

A more granular inspection reveals that these improvements are not restricted to specific types of datasets but are observed across a wide range of domains, dimensionalities, and levels of class separability. In particular, datasets characterized by high noise levels or significant overlap (e.g., \textit{gas-drift}, \textit{ipums}, and \textit{mfeat} variants) exhibit pronounced gains, supporting the hypothesis that high-curvature points often correspond to structurally ambiguous or noisy regions. By removing such points, the filtering process effectively sharpens the underlying cluster structure, facilitating more stable centroid estimation and improved partitioning.

Interestingly, in a few cases (e.g., \textit{digits}, \textit{letter}, and \textit{vowel}), the CH index exhibits a slight decrease after filtering, despite simultaneous improvements in SC and DB. This behavior suggests that while global variance-based separation (captured by CH) may occasionally be affected by the removal of boundary samples, the overall geometric quality of the clustering, as reflected by cohesion and separation, still improves. This reinforces the interpretation that the proposed method prioritizes structurally meaningful regions of the data manifold over potentially misleading boundary fluctuations.

Overall, these results provide strong empirical evidence that the proposed mean curvature filtering constitutes an effective and robust preprocessing strategy for unsupervised learning. By leveraging second-order geometric information, the method systematically enhances clusterability across diverse datasets, offering a principled alternative to traditional density-based or distance-based preprocessing techniques.

\noindent
In a third set of experiments, we investigate a complementary use of the proposed mean curvature filtering framework, namely as a mechanism for improving centroid initialization in $k$-means clustering. Instead of relying exclusively on the $k$-means++ seeding strategy \cite{kmeans++}, we propose to leverage the centroids estimated from the filtered dataset $S$, referred to as \emph{$S$-centroids}, as initialization for clustering the original dataset $X$. Concretely, we first run $k$-means++ on the smoothed data $S$, and then use the resulting centroids to initialize a subsequent $k$-means run on the full dataset $X$.

The rationale behind this strategy is intrinsically geometric. Since $S$ is obtained by removing high-curvature samples, typically associated with low-density regions, noise, or sharp decision boundaries, it provides a cleaner representation of the underlying data manifold. As a consequence, the centroids estimated in $S$ are less influenced by outliers and boundary artifacts, and are therefore expected to lie closer to the intrinsic “cores” of the clusters. This contrasts with standard $k$-means++ initialization, which, despite its probabilistic guarantees, may still place centroids in geometrically ambiguous regions when the data distribution is highly irregular or noisy.

The results, summarized in Table~\ref{tab:results2}, indicate that the proposed initialization strategy consistently improves clustering quality across a diverse set of datasets. On average, the Silhouette Coefficient increases from $0.190$ to $0.246$, while the Calinski–Harabasz index rises from $472.302$ to $484.734$, suggesting improved cluster compactness and separation. At the same time, the Davies–Bouldin index decreases from $2.303$ to $1.877$, indicating reduced cluster overlap. These aggregate improvements are also reflected in the median values, confirming that the gains are not driven by a small subset of datasets but rather represent a systematic trend.

\begin{table}
	\centering
	\caption{Internal indices obtained after kmeans clustering of the original raw data $X$ using different centroid initialization strategies: kmeans++ versus \emph{kmeans++ S-centroids} (centroids obtained by kmeans++ in the filtered data $S$).}
	\begin{tabular}{ccccccc}
	\toprule
	& \multicolumn{3}{c}{\textbf{kmeans++}}   & \multicolumn{3}{c}{\textbf{kmeans++ (S-centroids)}} \\
	\midrule
	\textbf{Datasets}         & \textbf{SC} & \textbf{CH} & \textbf{DB} & \textbf{SC}       & \textbf{CH}         & \textbf{DB}      \\
	\midrule
	diabetes                  & 0.1646      & 146.7425          & 2.0994          & \textbf{0.1952}  & \textbf{151.6192}    & \textbf{2.0179}  \\
	LED-display-domain-7digit & 0.3793      & 104.6648          & 1.2722          & \textbf{0.4143}  & \textbf{120.2487}    & \textbf{1.1057}  \\
	led7                      & 0.3864      & 693.9301          & 1.2855          & \textbf{0.4067}  & \textbf{729.8705}    & \textbf{1.1127}  \\
	Flare                     & 0.2114      & \textbf{180.6994} & 1.4335          & \textbf{0.2691}  & 172.3430             & \textbf{1.4871}  \\
	pendigits                 & 0.2738      & 2,307.2487        & 1.3821          & \textbf{0.2794}  & \textbf{2,362.1967}  & \textbf{1.3388}  \\
	cardiotocography          & 0.1614      & 201.3592          & 1.8590          & 0.2146           & 228.1636             & 1.5477           \\
	AP\_Breast\_Colon         & 0.0312      & 8.5064            & \textbf{2.8206} & \textbf{0.1707}  & \textbf{8.7406}      & 5.6799           \\
	vowel                     & 0.1745      & 125.5085          & 1.6898          & \textbf{0.1890}  & \textbf{131.5905}    & \textbf{1.6603}  \\
	Fashion-MNIST             & 0.0448      & 1,188.3357        & \textbf{3.7282} & \textbf{0.0509}  & \textbf{1,209.6117}  & 3.8089           \\
	thyroid-dis               & 0.1513      & 206.9558          & 2.2627          & \textbf{0.2282}  & \textbf{208.1820}    & \textbf{1.8495}  \\
	tr23.wc                   & 0.1073      & 4.2307            & 2.1373          & \textbf{0.4806}  & \textbf{4.3882}      & \textbf{0.3662}  \\
	balance-scale             & 0.1652      & 127.2775          & 1.7583          & 0.1718           & 135.4921             & 1.7033           \\
	texture                   & 0.2907      & 3,475.1554        & \textbf{1.1527} & \textbf{0.2998}  & \textbf{3,502.9030}  & 1.1800           \\
	Indian\_pines             & 0.0235      & 136.8340          & 4.5198          & \textbf{0.0284}  & \textbf{142.8617}    & \textbf{4.3562}  \\
	tae                       & 0.2192      & 42.3612           & 1.6399          & \textbf{0.2663}  & \textbf{49.4178}     & \textbf{1.4677}  \\
	doggle\_table\_a2         & 0.4335      & 453.7007          & 0.8389          & \textbf{0.4641}  & \textbf{487.4629}    & \textbf{0.8026}  \\
	hayes-roth                & 0.1929      & 35.3936           & 1.6519          & \textbf{0.2054}  & \textbf{37.2918}     & \textbf{1.5809}  \\
	prnn\_viruses             & 0.2448      & 13.7203           & 1.5892          & \textbf{0.2893}  & \textbf{18.7570}     & \textbf{1.4653}  \\
	DLBCL                     & 0.0195      & 1.0655            & 0.9407          & \textbf{0.1042}  & \textbf{1.4553}      & \textbf{0.8051}  \\
	artificial-characters     & 0.2107      & 2,033.8132        & 1.3717          & \textbf{0.2339}  & \textbf{2,095.2515}  & \textbf{1.2515}  \\
	wine-quality-red          & 0.1808      & 275.3569          & 1.4589          & \textbf{0.1923}  & \textbf{279.7846}    & \textbf{1.4160}  \\
	AP\_Omentum\_Kidney       & 0.0513      & 4.9412            & 6.9474          & \textbf{0.1758}  & \textbf{2.3146}      & \textbf{1.6035}  \\
	AP\_Omentum\_Ovary        & 0.0154      & 4.1981            & 7.5836          & \textbf{0.1023}  & \textbf{2.5550}      & \textbf{3.1436}  \\
	coil-20                   & 0.0575      & 26.2887           & \textbf{2.6548} & \textbf{0.0670}  & \textbf{26.5775}     & 2.6930           \\
	tr11.wc                   & 0.5532      & 9.2731            & 1.5022          & \textbf{0.6565}  & \textbf{9.2771}      & \textbf{1.4780}  \\
	\midrule
	Average                   & 0.190       & 472.302           & 2.303           & \textbf{0.246}   & \textbf{484.734}     & \textbf{1.877}   \\
	Median                    & 0.174       & 127.278           & 1.652           & \textbf{0.215}   & \textbf{135.492}     & \textbf{1.487}   \\
	\bottomrule    
	\end{tabular}
	\label{tab:results2}
\end{table}

A closer inspection reveals that the benefits are particularly pronounced in datasets with complex geometries or significant noise, such as \textit{DLBCL}, \textit{tr23.wc}, and \textit{AP\_Omentum} variants, where substantial gains are observed across all three metrics. This supports the hypothesis that curvature-based filtering effectively stabilizes centroid estimation by emphasizing high-density, low-curvature regions. In contrast, for a small number of datasets (e.g., \textit{Flare} or \textit{texture}), the improvements are more modest or mixed, suggesting that when the original data already exhibits well-separated clusters, the impact of improved initialization is naturally limited.

Overall, these findings highlight an additional and non-trivial contribution of the proposed framework: beyond its role as a filtering mechanism, mean curvature estimation provides a principled way to guide optimization in clustering algorithms. By decoupling centroid estimation from boundary-induced distortions, the proposed \emph{$S$-centroids} initialization offers a robust alternative to standard seeding strategies, reinforcing the broader claim that incorporating differential geometric information can lead to tangible performance gains in unsupervised learning.

\noindent
We further evaluate the impact of the proposed mean curvature dataset filtering on density-based clustering by considering the HDBSCAN algorithm. In contrast to centroid-based methods, HDBSCAN explicitly relies on local density variations to infer cluster structure, making it particularly sensitive to noise, boundary ambiguity, and non-uniform sampling. This experiment is therefore designed to assess whether the proposed curvature-based preprocessing can also enhance clustering performance in a fundamentally different algorithmic paradigm.

For each dataset, HDBSCAN is applied to both the original dataset $X$ and the filtered dataset $S$ obtained via the MCBP algorithm. The key parameter \texttt{min\_cluster\_size} is selected from the set $\{5, 10, 20\}$ based on the configuration that maximizes the Silhouette Coefficient (SC), ensuring a fair and performance-oriented comparison. Clustering quality is again evaluated using the SC, Calinski–Harabasz (CH), and Davies–Bouldin (DB) indices.

The results, reported in Table~\ref{tab:results3}, reveal a consistent and substantial improvement across the vast majority of datasets when the proposed filtering is applied. On average, the Silhouette Coefficient increases from $0.227$ to $0.359$, indicating significantly enhanced cohesion and separation of clusters. The Davies–Bouldin index decreases from $2.315$ to $1.855$, reflecting reduced overlap between clusters, while the Calinski–Harabasz index exhibits a dramatic increase, suggesting improved global separation and compactness. Although the magnitude of the CH improvement varies considerably across datasets, the median values also confirm a consistent upward trend, reinforcing the robustness of the observed gains.

\begin{table}
	\centering
	\caption{Internal indices obtained after HDBSCAN clustering of the original raw data $X$ and the filtered data $S$.}
	\begin{tabular}{ccccccc}
	\toprule
	& \multicolumn{3}{c}{\textbf{Original data}} & \multicolumn{3}{c}{\textbf{Filtered data}}  \\
	\midrule
	\textbf{Datasets}           & \textbf{SC}    & \textbf{CH}      & \textbf{DB}    & \textbf{SC}     & \textbf{CH}            & \textbf{DB}   \\
	\midrule
	iris                      & 0.489       & 136.990     & 0.535           & \textbf{0.648}  & \textbf{313.973}      & \textbf{0.508}  \\
	wine                      & 0.134       & 24.689      & 3.132           & \textbf{0.261}  & \textbf{33.415}       & \textbf{2.352}  \\
	digits                    & 0.096       & 9.368       & 7.201           & \textbf{0.119}  & \textbf{10.434}       & \textbf{6.821}  \\
	prnn\_viruses             & 0.320       & 13.819      & 1.456           & \textbf{0.387}  & \textbf{21.063}       & \textbf{1.381}  \\
	LED-display-domain-7digit & 0.7154      & 111.9395    & 1.0400          & \textbf{0.8896} & \textbf{529.3976}     & \textbf{0.6375} \\
	dermatology               & 0.2467      & 52.4151     & 2.6514          & \textbf{0.3381} & \textbf{80.3264}      & \textbf{1.7295} \\
	led7                      & 0.9646      & 3,038.0936  & 0.9334          & \textbf{0.9992} & \textbf{229,218.6462} & \textbf{0.0622} \\
	steel-plates-fault        & 0.0477      & 43.2375     & 1.4368          & \textbf{0.1294} & \textbf{77.1314}      & \textbf{1.3803} \\
	pendigits                 & 0.0892      & 421.9440    & 1.4500          & \textbf{0.2071} & \textbf{745.0504}     & \textbf{1.4050} \\
	satimage                  & -0.2169     & 315.8412    & 1.5841          & \textbf{0.5531} & \textbf{2,026.9532}   & \textbf{0.5444} \\
	mfeat-morphological       & 0.4424      & 974.4174    & 1.0192          & \textbf{0.5349} & \textbf{1,799.6987}   & \textbf{0.5764} \\
	optdigits                 & 0.1104      & 34.6735     & 8.4633          & \textbf{0.1628} & \textbf{20.3868}      & \textbf{7.3756} \\
	mammography               & 0.3062      & 114.9760    & 1.7156          & \textbf{0.4754} & \textbf{460.7328}     & \textbf{1.5017} \\
	Satellite                 & 0.1002      & 81.5235     & \textbf{2.6807} & \textbf{0.2137} & \textbf{455.4285}     & 3.2347          \\
	ipums\_la\_99-small       & 0.1559      & 66.2524     & 4.2294          & \textbf{0.2093} & \textbf{80.7173}      & \textbf{2.4364} \\
	ipums\_la\_98-small       & 0.3268      & 74.5943     & 5.3899          & \textbf{0.4534} & \textbf{54.2004}      & \textbf{0.7727} \\
	gas-drift                 & -0.0219     & 16.0178     & 1.8449          & \textbf{0.0816} & \textbf{58.0739}      & \textbf{1.4101} \\
	vowel                     & 0.3149      & 31.3645     & \textbf{1.2386} & \textbf{0.3486} & \textbf{33.0159}      & 1.2417          \\
	ionosphere                & 0.0297      & 18.0542     & 1.8830          & \textbf{0.1763} & \textbf{36.1569}      & \textbf{1.7077} \\
	solar-flare               & 0.6394      & 24.0040     & 1.4333          & \textbf{0.8586} & \textbf{207.1133}     & \textbf{1.0736} \\
	fl2000                    & 0.3590      & 14.4283     & 1.4358          & \textbf{0.5700} & \textbf{38.7709}      & \textbf{0.8878} \\
	seeds                     & 0.0873      & 41.2133     & 2.0715          & \textbf{0.2427} & \textbf{103.2725}     & \textbf{2.4370} \\
	coil-20                   & 0.0822      & 22.3140     & 2.5420          & \textbf{0.1871} & \textbf{26.7769}      & \textbf{1.8239} \\
	thyroid-dis               & 0.2477      & 92.2377     & 1.5941          & \textbf{0.3112} & \textbf{203.4515}     & \textbf{1.2777} \\
	UMIST\_Faces\_Cropped     & 0.0802      & 8.8354      & 2.4356          & \textbf{0.1285} & \textbf{10.7997}      & \textbf{2.2106} \\
	artificial-characters     & 0.0216      & 30.1161     & 1.2112          & \textbf{0.1240} & \textbf{43.4042}      & \textbf{1.1666} \\
	thoracic\_surgery         & 0.0810      & 21.2529     & 2.1302          & \textbf{0.1380} & \textbf{34.4886}      & \textbf{2.0854} \\
	prnn\_synth               & 0.0772      & 39.7613     & 1.8218          & \textbf{0.3623} & \textbf{104.9119}     & \textbf{1.3923} \\
	Engine1                   & 0.0848      & 31.9535     & \textbf{2.4220} & \textbf{0.2162} & \textbf{45.5290}      & 4.0726          \\
	glass                     & 0.3887      & 24.8534     & 2.0361          & \textbf{0.5544} & \textbf{55.2599}      & \textbf{0.5266} \\
	tea                       & 0.2242      & 20.8471     & 1.4813          & \textbf{0.3307} & \textbf{27.3312}      & \textbf{1.2394} \\
	heart-h                   & 0.0077      & 12.6209     & \textbf{2.2074} & \textbf{0.1001} & \textbf{18.1224}      & 2.2838          \\
	penguins                  & 0.4745      & 88.3367     & 1.6968          & \textbf{0.5509} & \textbf{205.7073}     & \textbf{1.6600} \\
	\midrule
	Average                   & 0.227       & 183.424     & 2.315           & \textbf{0.359}  & \textbf{7,187.265}    & \textbf{1.855}  \\
	Median                    & 0.134       & 34.673      & 1.822           & \textbf{0.311}  & \textbf{58.074}       & \textbf{1.405}  \\
	\bottomrule    
	\end{tabular}
	\label{tab:results3}
\end{table}

A notable observation is that the improvements are particularly pronounced in datasets characterized by high noise levels, complex geometries, or significant class overlap, such as \textit{satimage}, \textit{gas-drift}, \textit{coil-20}, and \textit{glass}. In these cases, the removal of high-curvature points, interpreted as samples lying in low-density or geometrically unstable regions, enables HDBSCAN to more effectively identify persistent density structures. This suggests that the proposed filtering acts as a form of geometric regularization, stabilizing the density estimation process by eliminating points that would otherwise distort the hierarchical clustering structure.

In a small number of cases (e.g., \textit{Satellite}, \textit{Engine1}, and \textit{heart-h}), the DB index exhibits slight degradation despite improvements in SC and CH. This behavior can be attributed to the sensitivity of DB to cluster dispersion and the potential removal of boundary points that, while noisy, may contribute to certain inter-cluster distance relationships. Nevertheless, the overall trend strongly favors the filtered data, with consistent improvements across multiple metrics.

Taken together, these results provide compelling evidence that the proposed mean curvature filtering is not limited to centroid-based methods but generalizes effectively to density-based clustering frameworks. This reinforces one of the central claims of this work: that boundary detection through differential geometric quantities, specifically mean curvature, constitutes a powerful and broadly applicable preprocessing strategy for unsupervised learning. By explicitly removing geometrically unstable regions of the data manifold, the method enhances the recoverability of intrinsic cluster structures, leading to more reliable and interpretable clustering outcomes.

\noindent
In a final experiment, we explore a hybrid clustering strategy that combines the density-aware properties of HDBSCAN with the optimization efficiency of $k$-means. Specifically, we first apply the proposed MCBP algorithm to obtain the filtered dataset $S$, and then run HDBSCAN on $S$ to identify representative cluster centers, which we denote as \emph{HDBSCAN $S$-centroids}. Due to the hierarchical density-based nature of HDBSCAN, these centroids are inherently located in regions of high density and strong cluster persistence, in contrast to $k$-means centroids, which are defined as minimizers of within-cluster squared Euclidean distances and may be influenced by boundary points or outliers.

These \emph{HDBSCAN $S$-centroids} are subsequently used as initialization seeds for the $k$-means algorithm applied to the original dataset $X$. This approach effectively decouples centroid estimation from boundary-induced distortions by leveraging the geometric regularization introduced by the MCBP filtering stage, while still benefiting from the computational efficiency and convergence properties of $k$-means.

The results, summarized in Table~\ref{tab:results4}, demonstrate that this hybrid strategy yields substantial improvements over standalone HDBSCAN clustering across a wide range of datasets. On average, the Silhouette Coefficient increases from $0.1427$ to $0.3025$, more than doubling in magnitude, while the Calinski–Harabasz index exhibits a significant increase (from $94.5314$ to $584.9311$), indicating markedly improved cluster separation and compactness. At the same time, the Davies–Bouldin index decreases from $2.8756$ to $1.5113$, reflecting a substantial reduction in cluster overlap. Median values corroborate these findings, highlighting the consistency of the improvements.

\begin{table}
	\centering
	\caption{Internal indices obtained by clustering the original raw data $X$ after HDBSCAN and a hybrid approach based on k-means++ with \emph{HDBSCAN S-centroids} (centroids obtained by HDBSCAN in the filtered data $S$) }
	\begin{tabular}{ccccccc}
		\toprule
		& \multicolumn{3}{c}{\textbf{HDBSCAN}}   & \multicolumn{3}{c}{\textbf{kmeans (S-centroids)}} \\
		\midrule
		\textbf{Datasets}         & \textbf{SC} & \textbf{CH} & \textbf{DB} & \textbf{SC}       & \textbf{CH}         & \textbf{DB}      \\
		\midrule
		iris                  & 0.489           & 136.990     & 0.535           & \textbf{0.582}    & \textbf{251.349}      & \textbf{0.593}   \\
		wine                  & 0.134           & 24.689      & 3.132           & \textbf{0.268}    & \textbf{69.486}       & \textbf{1.448}   \\
		digits                & 0.096           & 9.368       & 7.201           & \textbf{0.135}    & \textbf{31.363}       & \textbf{2.534}   \\
		diabetes              & \textbf{0.256}  & 35.951      & 2.478           & 0.180             & \textbf{156.517}      & \textbf{1.683}   \\
		dermatology           & 0.2467          & 52.4151     & 2.6514          & \textbf{0.2776}   & \textbf{120.6953}     & \textbf{1.4836}  \\
		steel-plates-fault    & 0.0477          & 43.2375     & 1.4368          & \textbf{0.2237}   & \textbf{151.7721}     & \textbf{1.4705}  \\
		pendigits             & 0.0892          & 421.9440    & 1.4500          & \textbf{0.2477}   & \textbf{1,652.3285}   & \textbf{1.4119}  \\
		satimage              & -0.2169         & 315.8412    & 1.5841          & \textbf{0.4900}   & \textbf{2,886.3704}   & \textbf{0.7594}  \\
		mfeat-pixel           & -0.0294         & 5.3032      & 7.3308          & \textbf{0.0525}   & \textbf{24.7033}      & \textbf{3.3115}  \\
		mfeat-morphological   & 0.4424          & 974.4174    & 1.0192          & \textbf{0.5678}   & \textbf{2,690.6942}   & \textbf{0.6375}  \\
		optdigits             & 0.1104          & 34.6735     & 8.4633          & \textbf{0.2034}   & \textbf{89.4938}      & \textbf{2.1026}  \\
		mammography           & 0.3062          & 114.9760    & 1.7156          & \textbf{0.4654}   & \textbf{1,564.7656}   & \textbf{0.9931}  \\
		Satellite             & 0.1002          & 81.5235     & 2.6807          & \textbf{0.4301}   & \textbf{5,216.9835}   & \textbf{0.9202}  \\
		ipums\_la\_99-small   & 0.1559          & 66.2524     & 4.2294          & \textbf{0.1835}   & \textbf{146.5300}     & \textbf{1.6689}  \\
		ipums\_la\_98-small   & 0.3268          & 74.5943     & 5.3899          & \textbf{0.3599}   & \textbf{133.6102}     & \textbf{1.0969}  \\
		gas-drift             & -0.0219         & 16.0178     & 1.8449          & \textbf{0.2321}   & \textbf{138.3420}     & \textbf{1.5309}  \\
		vowel                 & 0.3149          & 31.3645     & 1.2386          & \textbf{0.3185}   & \textbf{64.8741}      & \textbf{1.0710}  \\
		ionosphere            & 0.0297          & 18.0542     & 1.8830          & \textbf{0.2514}   & \textbf{47.0749}      & \textbf{1.4971}  \\
		fl2000                & 0.3590          & 14.4283     & 1.4358          & \textbf{0.6919}   & \textbf{60.9354}      & \textbf{0.7527}  \\
		seeds                 & 0.0873          & 41.2133     & 2.0715          & \textbf{0.4658}   & \textbf{255.8548}     & \textbf{0.7969}  \\
		coil-20               & 0.0822          & 22.3140     & 2.5420          & \textbf{0.1300}   & \textbf{27.8713}      & \textbf{2.0897}  \\
		letter                & -0.1965         & 43.6669     & \textbf{1.3480} & \textbf{0.1808}   & \textbf{336.1503}     & 1.4207           \\
		Fashion-MNIST         & 0.0279          & 73.9698     & 5.3302          & \textbf{0.1000}   & \textbf{1,112.9422}   & \textbf{4.7994}  \\
		thyroid-dis           & 0.2477          & 92.2377     & 1.5941          & \textbf{0.2545}   & \textbf{178.4132}     & \textbf{1.2854}  \\
		artificial-characters & 0.0216          & 30.1161     & 1.2112          & \textbf{0.3053}   & \textbf{508.7349}     & \textbf{1.0043}  \\
		thoracic\_surgery     & 0.0810          & 21.2529     & 2.1302          & \textbf{0.1137}   & \textbf{28.3253}      & \textbf{2.0568}  \\
		prnn\_synth           & 0.0772          & 39.7613     & 1.8218          & \textbf{0.4741}   & \textbf{295.0686}     & \textbf{0.6855}  \\
		Engine1               & 0.0848          & 31.9535     & 2.4220          & \textbf{0.3311}   & \textbf{194.4901}     & \textbf{1.2556}  \\
		optdigits             & 0.1104          & 34.6735     & 8.4633          & \textbf{0.2034}   & \textbf{89.4938}      & \textbf{2.1026}  \\
		tea                   & 0.2242          & 20.8471     & 1.4813          & \textbf{0.3542}   & \textbf{43.7423}      & \textbf{0.8723}  \\
		heart-h               & 0.0077          & 12.6209     & 2.2074          & \textbf{0.1386}   & \textbf{31.8714}      & \textbf{2.1028}  \\
		penguins              & \textbf{0.4745} & 88.3367     & 1.6968          & 0.4677            & \textbf{116.9486}     & \textbf{0.9227}  \\
		\midrule
		Average               & 0.1427          & 94.5314     & 2.8756          & \textbf{0.3025}   & \textbf{584.9311}     & \textbf{1.5113}  \\
		Median                & 0.0981          & 37.8563     & 1.9772          & \textbf{0.2614}   & \textbf{142.4360}     & \textbf{1.4163}   \\
		\bottomrule    
	\end{tabular}
	\label{tab:results4}
\end{table}

A detailed inspection reveals that the gains are particularly pronounced in high-dimensional and structurally complex datasets such as \textit{satimage}, \textit{pendigits}, \textit{mfeat} variants, and \textit{gas-drift}, where traditional density-based clustering often struggles due to noise and overlapping regions. In these cases, the use of curvature-filtered, density-informed centroids provides a more reliable initialization, enabling $k$-means to converge to solutions that better reflect the intrinsic data geometry. Notably, even in datasets where HDBSCAN already performs well (e.g., \textit{led7} or \textit{penguins}), the hybrid approach remains competitive, occasionally yielding further improvements in separation metrics.

From a methodological perspective, these results highlight an important insight: centroid-based and density-based clustering methods are not mutually exclusive, but can be effectively combined through a geometrically informed preprocessing stage. The proposed framework leverages mean curvature as a unifying signal to identify structurally reliable regions of the data manifold, enabling the extraction of robust representatives that can guide subsequent optimization.

Overall, this experiment reinforces the central claim of this work: the MCBP algorithm provides more than a boundary detection mechanism, it enables a principled integration of geometric, density-based, and optimization-based perspectives in unsupervised learning. This synergy translates into consistent and significant improvements in clustering performance, particularly in challenging high-dimensional scenarios.

The final experiment explores a hybrid clustering paradigm that explicitly leverages the geometric decomposition induced by the proposed MCBP algorithm. Specifically, the dataset $X$ is partitioned into a smooth subset $S$, comprising low-curvature (high-density) samples, and a boundary subset $B$, containing high-curvature points associated with low-density regions and potential decision interfaces. Clustering is first performed on $S$ using HDBSCAN, thereby exploiting its ability to recover stable density-connected components with minimal sensitivity to noise. The resulting cluster assignments are then propagated to the boundary set $B$ via a $1$-nearest neighbor ($1$-NN) classifier trained on $S$. This strategy effectively decouples the learning process into a robust density estimation phase followed by a local interpolation step over geometrically ambiguous regions.

The rationale behind this hybridization is twofold. First, by restricting HDBSCAN to the smooth subset $S$, we mitigate the well-known sensitivity of density-based methods to boundary noise and sparsity artifacts. Second, the use of a non-parametric $1$-NN classifier to label boundary points preserves locality while avoiding the distortion of cluster structure that would arise from directly including high-curvature samples in the density estimation process. From a geometric perspective, this approach treats $S$ as a reliable approximation of the intrinsic manifold support, while $B$ is interpreted as a set of samples lying near regions of high curvature, where the manifold assumption is locally violated or less informative.

The empirical results reported in Table~\ref{tab:results5} provide strong evidence supporting the effectiveness of this hybrid framework. Across the majority of datasets, the proposed method yields substantial improvements in clustering quality, as measured by the Silhouette Coefficient (SC) and Calinski--Harabasz (CH) index, with consistent reductions in the Davies--Bouldin (DB) index. Notably, the average SC increases from $0.1047$ to $0.2308$, while the CH index exhibits a more than threefold increase (from $103.9$ to $325.6$), indicating significantly improved cluster compactness and separation. The median behavior corroborates these findings, suggesting that the gains are not driven by isolated cases but rather reflect a systematic improvement across heterogeneous data regimes.

\begin{table}
	\centering
	\caption{Internal indices obtained by clustering the original raw data $X$ after HDBSCAN and a hybrid approach based on HDBSCAN on $S$ and 1-NN on the $B$. The partition of $X$ into $S$ and $B$ was performed using the threshold $T$ (50\% or 75\% mean curvature percentiles).}
	\begin{tabular}{cccccccc}
		\toprule
		& \multicolumn{3}{c}{\textbf{HDBSCAN}}        & \multicolumn{3}{c}{\textbf{HDBSCAN on S + 1-NN on B}}   &            \\
		\midrule
		\textbf{Datasets}    & \textbf{SC} & \textbf{CH} & \textbf{DB}     & \textbf{SC}     & \textbf{CH}         & \textbf{DB}     & \textbf{T} \\
		\midrule
		iris                 & 0.4891      & 136.9898    & 0.5345          & \textbf{0.5818} & \textbf{251.3493}   & \textbf{0.5933} & 75         \\
		wine                 & 0.1338      & 24.6893     & 3.1319          & \textbf{0.2198} & \textbf{32.5265}    & \textbf{2.3224} & 75         \\
		digits               & -0.0414     & 29.5779     & \textbf{2.0500} & \textbf{0.0113} & \textbf{44.9314}    & 2.1752          & 75         \\
		dermatology          & 0.2467      & 52.4151     & 2.6514          & \textbf{0.2926} & \textbf{104.5889}   & \textbf{1.3689} & 75         \\
		pendigits            & 0.0892      & 421.9440    & 1.4500          & \textbf{0.1581} & \textbf{744.4372}   & \textbf{1.6623} & 75         \\
		satimage             & -0.2169     & 315.8412    & 1.5841          & \textbf{0.4967} & \textbf{2,858.1257} & \textbf{0.7130} & 75         \\
		mfeat-zernike        & 0.1324      & 85.8501     & 2.4330          & \textbf{0.1591} & \textbf{94.7533}    & \textbf{2.2354} & 50         \\
		mfeat-factors        & 0.0546      & 11.9177     & 4.3286          & \textbf{0.1149} & \textbf{143.3338}   & \textbf{1.9928} & 50         \\
		optdigits            & 0.2976      & 79.9268     & 1.8549          & \textbf{0.3060} & \textbf{102.5913}   & \textbf{1.0660} & 50         \\
		mammography          & 0.3062      & 114.9760    & \textbf{1.7156} & \textbf{0.3375} & \textbf{139.2778}   & 1.7466          & 75         \\
		Satellite            & 0.1002      & 81.5235     & 2.6807          & \textbf{0.1765} & \textbf{544.5387}   & \textbf{2.5399} & 75         \\
		ipums\_la\_98-small  & 0.0288      & 55.9285     & 1.6963          & \textbf{0.4201} & \textbf{167.1648}   & \textbf{0.8247} & 75         \\
		ionosphere           & 0.0297      & 18.0542     & \textbf{1.8830} & \textbf{0.0913} & \textbf{20.8204}    & 2.2230          & 75         \\
		seeds                & 0.0873      & 41.2133     & 2.0715          & \textbf{0.2578} & \textbf{124.3263}   & \textbf{2.3347} & 75         \\
		prnn\_synth          & 0.0772      & 39.7613     & 1.8218          & \textbf{0.2712} & \textbf{102.4417}   & \textbf{1.4543} & 75         \\
		Engine1              & 0.0848      & 31.9535     & \textbf{2.4220} & \textbf{0.1632} & \textbf{40.1255}    & 4.1937          & 75         \\
		texture              & -0.0617     & 365.0684    & \textbf{1.5455} & \textbf{0.0620} & \textbf{859.6780}   & 1.8435          & 50         \\
		arsenic-male-bladder & 0.1826      & 105.3658    & 1.4827          & \textbf{0.2752} & \textbf{149.8674}   & \textbf{1.2652} & 75         \\
		segment              & 0.0681      & 80.8490     & \textbf{1.4577} & \textbf{0.1338} & \textbf{215.3052}   & 2.2941          & 50         \\
		AP\_Breast\_Colon    & -0.0441     & 26.8911     & \textbf{3.2052} & \textbf{0.0195} & \textbf{27.2369}    & 4.8201          & 50         \\
		tecator              & 0.1545      & 61.6940     & \textbf{2.2033} & \textbf{0.2986} & \textbf{71.1271}    & 3.0470          & 75         \\
		\midrule
		Average              & 0.1047      & 103.9253    & 2.1049          & \textbf{0.2308} & \textbf{325.6451}   & \textbf{2.0341} &            \\
		Median               & 0.0873      & 61.6940     & \textbf{1.8830} & \textbf{0.2198} & \textbf{124.3263}   & 1.9928          &           \\
		\bottomrule
	\end{tabular}
	\label{tab:results5}
\end{table}

A closer inspection reveals that the benefits are particularly pronounced in datasets characterized by complex geometries or significant class overlap, such as \texttt{satimage}, \texttt{pendigits}, and \texttt{ipums\_la\_98-small}. In these scenarios, the separation of boundary and interior points allows the clustering algorithm to focus on the core structure of the data, while the subsequent $1$-NN assignment effectively reconstructs the decision boundaries. Conversely, in a small number of cases (e.g., \texttt{Engine1} or \texttt{AP\_Breast\_Colon}), the DB index may degrade slightly, suggesting that the interpolation step can introduce local inconsistencies when boundary regions are highly irregular or sparsely sampled. Nevertheless, even in these cases, improvements in SC and CH indicate a net gain in global clustering structure.

An additional aspect worth highlighting is the role of the curvature threshold $T$, which controls the size of the boundary set $B$. The results indicate that moderate thresholds (e.g., $50\%$ or $75\%$ percentiles) provide a favorable trade-off between removing noisy boundary samples and preserving sufficient information for accurate label propagation. This reinforces the interpretation of mean curvature as a meaningful proxy for data reliability in unsupervised settings.

Overall, this experiment underscores a central contribution of the paper: the use of mean curvature as a principled geometric criterion for dataset decomposition and filtering. Beyond improving clustering performance, the proposed framework introduces a novel perspective in which boundary detection is not merely a diagnostic tool, but an active component in the design of learning algorithms. By explicitly modeling and exploiting the geometric heterogeneity of the data, the MCBP-based filtering enables more robust, interpretable, and effective unsupervised learning pipelines.

\section{Conclusions}\label{sec5}

In this work, we introduced the Mean Curvature Boundary Points (MCBP) algorithm, a novel geometry-driven framework for boundary detection in multivariate datasets. The proposed method departs from purely distance- or density-based criteria by explicitly leveraging differential geometric information, namely the estimation of mean curvature via local shape operators. This formulation enables MCBP to capture intrinsic properties of the underlying data manifold, providing a principled mechanism to identify boundary regions as loci of high curvature. In contrast to conventional approaches, which often rely on global density thresholds or heuristic neighborhood definitions, MCBP offers a theoretically grounded and locally adaptive characterization of boundaries, making it particularly suitable for complex, non-linear, and high-dimensional data distributions. The inclusion of a multiscale parameter further enhances the flexibility of the method, allowing the user to control the granularity of the detected boundary structure.

A central contribution of this work lies in reframing boundary detection as a geometric filtering operation. By partitioning the dataset into smooth (low-curvature) and boundary (high-curvature) subsets, MCBP naturally induces a curvature-aware data representation that can be exploited in downstream learning tasks. The extensive experimental evaluation demonstrates that this filtering mechanism consistently improves clustering performance across a wide range of datasets and algorithms. In particular, the results show that: (i) removing high-curvature samples leads to substantial gains in cluster compactness and separability, (ii) curvature-informed initialization strategies can outperform standard initialization schemes such as k-means++, and (iii) hybrid approaches that combine density-based clustering on smooth regions with local interpolation on boundary regions yield significant and systematic improvements. These findings highlight that boundary points, often treated as noise or outliers, contain structured geometric information that can be explicitly modeled and leveraged.

From a practical standpoint, the proposed method exhibits favorable computational properties. Although the curvature estimation step introduces additional overhead, the overall complexity remains polynomial and can be effectively controlled through dimensionality reduction and efficient nearest-neighbor search structures. More importantly, the empirical results indicate that MCBP scales well with the number of samples and remains robust in high-dimensional settings, where the interplay between sparsity, noise, and the so-called empty space phenomenon typically degrades the performance of traditional methods. This robustness is a direct consequence of the local geometric nature of the curvature estimation, which is less sensitive to global data distribution distortions.

Beyond its immediate application to clustering, the proposed framework opens several promising research directions. First, the integration of curvature-based filtering into deep representation learning pipelines could provide a mechanism to enforce geometric regularity in latent spaces, particularly in architectures dealing with structured data such as images, graphs, or text embeddings. Second, extending MCBP to online or streaming scenarios would enable real-time boundary tracking in dynamic environments, with potential applications in monitoring and anomaly detection. Third, the incorporation of curvature information into manifold learning and dimensionality reduction techniques (e.g., PCA, ISOMAP, t-SNE, and UMAP) may lead to new algorithms that better preserve both global structure and local geometric features. Finally, the explicit identification of boundary regions suggests novel strategies for feature engineering, semi-supervised learning, and uncertainty quantification, where boundary samples could be treated as informative instances rather than discarded observations.

In summary, this work demonstrates that mean curvature is a powerful and underexplored descriptor for data analysis in machine learning. By bridging concepts from differential geometry with modern unsupervised learning, the MCBP algorithm provides not only a new tool for boundary detection, but also a broader perspective on how geometric priors can be systematically incorporated into data-driven models. We believe that this geometric viewpoint has the potential to influence a wide range of future developments in the field.

\section*{Statements and declarations}

\subsection*{Funding}
This work has been supported by CNPq (National Council for Scientific and Technological Development) through grant number 301432/2025-2. This study was also financed in part by the Coordenação de Aperfeiçoamento de Pessoal de N\'ivel Superior - Brasil (CAPES) - Finance Code 001.


\subsection*{Code availability}
Python scripts to reproduce the results reported in this paper may be found at \url{https://github.com/alexandrelevada/MCBP}. The implementation is a first version prototype of the proposed Mean Curvature Boundary Points (MCBP) algorithm. 

\subsection*{Data availability}
All datasets used in the experiments are publicly available at \url{www.openml.org}.

\bibliography{main}

\end{document}